\documentclass{article}

\usepackage{arxiv}

\usepackage[utf8]{inputenc} 
\usepackage[T1]{fontenc}    
\usepackage{hyperref}       
\usepackage{url}            
\usepackage{booktabs}       
\usepackage{amsfonts}       
\usepackage{nicefrac}       
\usepackage{microtype}      
\usepackage{cleveref}       
\usepackage{multirow}        
\usepackage{lipsum}         
\usepackage{graphicx}
\usepackage{natbib}
\usepackage{doi}
\usepackage{rotating}

\title{Artificial Intelligence and Innovation Ecosystem: Evolutionary Developments, Challenges, and Future Directions}

\newif\ifuniqueAffiliation
\uniqueAffiliationtrue

\usepackage{authblk}

\newbox{\orcid}\sbox{\orcid}{\includegraphics[scale=0.06]{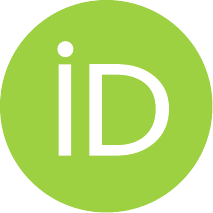}} 
\author[1]{%
	\href{https://orcid.org/0000-0002-9065-8724}{\usebox{\orcid}\hspace{1mm}Zhimin Zhang \thanks{\texttt{Corresponding author (zhangzhimin@uest.edu.gr).}}}%
}
\author[2,1]{Chengzhen Ma}

\author[1]{Jia Chai}

\author[1]{Rongxin Zhan}

\author[3,4]{Huansheng Ning}

\author[3]{Lingfeng Mao}

\author[1]{Dan Zhang}

\author[1]{Suiping Jiang}

\affil[1]{Beijing Institute of Computer Technology and Applications, Beijing 100854, China}

\affil[2]{School of Cyberspace Science and Technology, Beijing Institute of Technology, Beijing 100081, China}

\affil[3]{School of Computer and Communication Engineering, University of Science and Technology Beijing, Beijing 100083, China}

\affil[4]{Beijing Engineering Research Center for Cyberspace Data Analysis and Applications, Beijing 100083, China}

\renewcommand{\shorttitle}{Artificial Intelligence and Innovation Ecosystem}

\hypersetup{
pdftitle={Artificial Intelligence and Innovation Ecosystem: Evolutionary Developments, Challenges, and Future Directions},
pdfsubject={q-bio.NC, q-bio.QM},
pdfauthor={Zhimin Zhang, et al.},
pdfkeywords={Artificial Intelligence, Innovation Ecosystem, Innovative Evolutionary Development, Innovation Network, Value Co-Creation},
}

\begin{document}
\maketitle

\begin{abstract}
	The development of the Innovative Ecosystem (IE) presents a new paradigm for economic integration, collaborative advancement, and shared achievements. The rise of Artificial Intelligence (AI) has significantly accelerated the global processes of digitization, informatization, and intelligence. Exploring how AI can leverage inherent characteristics to influence the development trajectory of IE is a topic that warrants further investigation. Given AI's increasing prominence and role within IE, the paper analyzes this new form, examining both AI's unique contributions to IE and its potential challenges. Firstly, the paper synthesizes the conceptual frameworks surrounding IE, decomposing them into manifestations in physical, social, and thinking spaces. Furthermore, the concept of Artificial Intelligence IE (AIIE) is introduced from a spatial perspective, with an exploration of the characteristics AI contributes to IE. Subsequently, the paper employs an evolutionary perspective to analyze the roles provided by AI during different development periods of AIIE. The paper then verifies the feasibility, effectiveness, and rationality of the AIIE's definition and analyzes AIIE development from an evolutionary perspective using enterprise development examples. Finally, acknowledging AI's inherent limitations, the paper examines potential challenges facing AIIE in the future from four perspectives, aiming to identify new research avenues for the further development of AIIE.
\end{abstract}

\keywords{Artificial Intelligence \and Innovation Ecosystem \and Innovative Evolutionary Development \and Innovation Network \and Value Co-Creation}

\section{Introduction}

Innovation Ecosystem (IE) is an advanced network that includes participants such as governments, academic institutions, and innovation companies, as well as external environments like culture, law, and education, all aimed at fostering technological and business model innovation. 
These participants interact with the external environment, influencing each other and facilitating innovation through resource sharing, knowledge exchange, and collaborative cooperation (\cite{mercier2020platform}). 
As globalization deepens, IE has significantly contributed to sustainable global development. 
Furthermore, the ongoing evolution of IE has led many participants to recognize that, in light of current trends, a balance of cooperation and competition can drive mutually beneficial innovation and value enhancement.

The advancement of information technology, particularly Artificial Intelligence (AI), has accelerated the global processes of digitization, virtualization, and intelligence, fostering the ubiquitous connection and deep integration of physical space, social space, thinking space, and cyberspace (\cite{ning2018general}).
From preset rules and logical reasoning to powerful computing and self-learning capabilities, AI has not only enhanced the automation of digital processes but also profoundly impacted fields such as healthcare, education, finance, and transportation. 
Ongoing attention and exploration in AI will continue to drive technological breakthroughs and expand applications. 
Additionally, AI has improved cross-domain collaboration, optimized supply chains for demand allocation, and enhanced international interactions. 
Consequently, the interplay between AI and globalization has significantly contributed to economic integration and market expansion.

The unprecedented progress and changes brought by AI have laid a strong foundation for IE's digital transformation, driving its rapid growth and facilitating collaborative development. 
As AI's role within IE continues to expand, participants' decisions will increasingly be shaped by AI, potentially altering the dominant positions of both. 
However, if this technology is not effectively utilized or controlled, IE may fail to develop as envisioned. 
Consequently, some studies have begun to explore AI's role in IE, particularly its remarkable contributions to assisted decision-making, cross-domain collaboration, and value co-creation.

Studies on AI-assisted IE are still in their infancy, lacking a unified, recognized, and comprehensive definition, and the existing literature on this topic is also inconsistent (\cite{roundy2024understanding}). 
This ambiguity hampers efforts to identify the emerging IE's characteristics and effectively describe how it uniquely contributes to sustainable global development in the rapidly iterating and evolving era of AI. 
Moreover, current research often focuses on the impact and role of AI on IE with a certain development scale from the perspectives of participant (including external environment) composition, resource sharing, and innovative cooperation.
However, IE at different developmental periods varies in basic conditions, environmental factors, task orientations, and value goals. 
As a result, current studies tend to overlook the dynamic changes in AI demand throughout different periods of IE development.

Given the current lack of detailed descriptions, this paper breaks down the existing descriptions of IE into manifestations in physical, social, and thinking spaces. Based on the impact of AI-centered cyberspace in these three spaces, a conceptual description of AIIE is summarized, and its unique characteristics are analyzed.
Due to the insufficient review of AI's dynamic impact on the development of IE (\cite{matt2021role}), the paper categorizes the development periods of AIIE into startup, growth, and maturity from an evolutionary perspective.
It explores, elaborates, and summarizes the unique roles and requirements of AI at each period. 
Subsequently, drawing on the development experience of the medical company Owkin, the paper empirically demonstrates the rationality of the AIIE concept's proposal and analysis and its evolutionary development.
Additionally, the paper acknowledges technological development's dual impact and further examines the potential challenges and risks that AI poses to IE development. 
The main contributions of the paper include:

\begin{itemize}
  \item To address the insufficient conceptualization of AI and IE integration in existing studies, the paper breaks down the existing descriptions of IE from the perspectives of physical, social, and thinking spaces. After analyzing the impact of AI-dominated cyberspace on the three spaces, a conceptual description of AIIE is proposed and compared with similar research from multiple perspectives. Moreover, based on the unique performance of AI in seven aspects, the paper summarises and analyses the prominent characteristics of AIIE in comparison to traditional IE.
  \item In response to the current lack of analysis regarding AI's specific role at different AIIE development periods, the paper adopts an evolutionary perspective to examine the development patterns and demand characteristics of AIIE across the startup, growth, and maturity periods, highlighting the key roles that AI provides and the potential impacts in entails.
  \item To enhance understanding, verify feasibility, and demonstrate the application value, the paper summarises the development of the medical company Owkin as an example to validate AIIE. On the one hand, the paper analyzes the rationality of the AIIE concept from the perspectives of participant compositions, cooperative-competitive relationships, and ultimate goals. On the other hand, the paper demonstrates the feasibility of describing the unique role of AI in AIIE development's different periods by examining the changes in Owkin's business focus during the development process.
  \item As an answer to technological development's dual impact, the paper summarizes the potential challenges and risks that AI poses to AIIE development in terms of interpretability, energy conservation, fairness, and harmony, and analyzes future development directions and opportunities for AIIE in these four areas.
\end{itemize}

The paper is organized as follows. 
In Section 2, a summary of concepts related to IE from a spatial perspective is presented, followed by an analysis of the differences between AIIE and traditional IE from seven perspectives after refining AIIE's concept. 
Section 3 describes the unique roles and capabilities that AI contributes to AIIE development at various periods, including the control adaptation ability and unique creative ability in the startup period, the cross-domain communication ability and the collaborative decision-making ability in the growth period, and the decentralized ability and the privacy security protection ability in the maturity period. 
Based on Owkin's development experience, Section 4 illustrates the feasibility, effectiveness, and rationality of the AIIE's conceptual definition and development focus at different periods through case analysis.
Section 5 addresses the potential challenges and risks that AI poses to AIIE, focusing on interpretability, energy conservation, fairness, and harmony, and analyzes future development trends and opportunities in these areas. 
Section 6 concludes the entire paper with a summary of key points.

\section{Definition and Characteristics of AIIE}

\subsection{Definition of AIIE}

IE is analogous to a natural ecosystem, highlighting the interdependence of various participants who cooperate, share, and communicate to adapt to technological advancements, market changes, and policy adjustments through evolution. 
In such interdependent relationships, cooperation and healthy competition are crucial, guided by ultimate goals of achievement.
Influenced by factors such as field, scenario, and scope, IE has given rise to several related concepts, including business ecosystem (\cite{adner2006match}), industry 4.0 IE (\cite{matt2021role,benitez2020industry}), and smart product IE (\cite{kahle2020smart}).

AI is crucial for data analysis, predictive modeling, and operational optimization, serving as a powerful driving force of digital transformation. 
Under the influence of AI technology, system-level concepts such as Cyber-Physical-Social Systems (CPSS) (\cite{zeng2020survey}), General Cyberspace (GC) (\cite{ning2018general}), and PhiNet of Things (PoT) (\cite{ning2020phinet}) have been successively proposed to emphasize the optimization and improvement of traditional systems through enhanced productivity. A comprehensive analysis of these concepts underscores the profound impact of cyberspace, as represented by AI, on traditional physical, social, and even thinking spaces. Physical space is the foundation of spatial factors, mainly composed of spatial entities (such as people and objects) and corresponding environments. Social space considers the social attributes, behaviors, and interrelationships of entities. Consequently, thinking space primarily processes related brain activities or observes reactions in the realm of thinking.

\begin{table}
\caption{Summary of IE's Definition and Description from Physical, Social, and Thinking Spaces}
{
  \renewcommand{\arraystretch}{1.5}
  \begin{tabular}{p{3cm}p{4.5cm}p{3.5cm}p{3.8cm}} \toprule
    & \multicolumn{3}{c}{Definition and description} \\ \cmidrule{2-4}
    Concept & Physical space & Social space & Thinking space \\ \midrule
    IE  (\cite{mercier2020platform}) & Companies, research institutes, governments, educational institutions, technology, and startups/small/medium-sized enterprises & Interaction and influence & - \\
    Smart product IE (\cite{kahle2020smart}) & Companies with different complementary technological capabilities & Integration through collaborative development efforts & - \\
    Manufacturing IE  (\cite{reynolds2018strengthening}) & Participants and entities & Collaboration and combination respective resources & Foster technological development and innovation \\
    Industry 4.0 IE  (\cite{benitez2020industry}) & Digital technologies, information systems, and processing technologies & Interconnection & A high degree of capability interdependence and technological complementarity \\
    Business IE  (\cite{paasi2023modeling}) & Organizations and individuals & Interaction & The organic foundation of the business world \\ \bottomrule
  \end{tabular}
}
\label{Table: Summary of IE's Definition and Description from Physical, Social, and Thinking Spaces}
\end{table}

As a manifestation of the ecosystem, IE can also be represented in three spaces and will be influenced by AI-oriented cyberspace. For IE, its manifestation in physical space is compositions of participants; in social space, it is the manifestation of cooperative-competitive relationships; and in thinking space, it is manifested as ultimate goals. Based on this classification, the paper decomposes the existing concepts of IE and summarizes them in Table \ref{Table: Summary of IE's Definition and Description from Physical, Social, and Thinking Spaces}.

\subsubsection{Compositions of Participants in Physical Space}

Participants define the composition and scale of IE, representing IE's concrete manifestation in the physical space. 
Startups and technology companies with project experience are vital components that enhance industry competitiveness and facilitate information resource exchange. 
To boost innovation capabilities, industry research institutes, educational institutions, and higher education entities with cutting-edge knowledge and an exploratory spirit invigorate the development of IE.

Given AI's disruptive and uncertain nature, effective regulation is essential for maintaining IE's vitality and healthy operation. 
Governments and decision-making bodies play a crucial role in providing such regulatory support. 
Compared to mandatory regulation, dynamic and sandbox regulatory ways offer greater flexibility for AI development within IE (\cite{fenwick2018business}).

To ensure the successful execution of key activities like innovative research and technological development, venture capital, private equity, and angel investments serve as important funding sources. 
Additionally, incubators can help reduce innovation costs, facilitate achievement transformation, and enhance system efficiency by providing venue, infrastructure, and service support.

Finally, target users are also a critical component of IE. 
Their experiences and feedback can significantly influence IE's future direction, aiding the ecosystem in adjusting and making decisions that align with market and social development needs.

With the addition of AI, this technology can act as a decision-maker or technical advisor, assisting in various decision-making processes rather than being limited to human participants in the physical space. 
By leveraging AI technology, IE can process and analyze vast amounts of data, providing valuable insights in areas such as development trends, resource allocation, and personalized recommendations. 
Combining these suggestions, participants such as governments and institutions can make more informed and comprehensive development decisions.

With the advancement of Generative AI (GAI), large language models like OpenAI ChatGPT, Amazon Lex, and Google Dialogflow can perform complex interactive functions, including creative writing, answering questions, and content generation, further enhancing innovation capabilities within the ecosystem.

\subsubsection{Cooperative-Competitive Relationships in Social Space}

The second element in describing IE is cooperative-competitive relationships, which reflect its attributes in the social space. 
Achieving systemic development is challenging for any single participant. 
Thus, cooperation is crucial for fostering innovation and is a key focus of current research on interdependence. 
Participants can achieve high capability dependence and technological complementarity through knowledge sharing, resource integration, and information exchange, which helps reduce duplicate investments and lower costs (\cite{benitez2020industry}).

Moreover, healthy competition enhances the capabilities of IE. 
Different participants may compete for resources, technology, and users within similar market environments, which is conducive to driving ecosystem vitality, improving work efficiency, and stimulating innovation while enhancing product and service quality. 
Competition is relative, and competitors may also find opportunities for cooperation to address industry challenges, such as standardization and rule-making jointly.

In both cooperation and competition, it is essential for participants to maintain positive relationships and avoid actions that undermine the rights of others or lead to impulsive decisions (\cite{kahle2020smart,nambisan2013entrepreneurship}). 
Additionally, regulatory measures, including policies, regulations, and ordinances, can significantly impact participant relationships. 
For instance, AI development policies may encourage cross-domain collaboration, while antitrust laws are designed to protect healthy competition.

In terms of cooperative-competitive relationships, the application of AI can eliminate geographical limitations and industry barriers, fostering deeper collaboration among participants from various fields in the social space. 
Additionally, the development of General AI (AGI) has endowed this technology with the ability to perform diverse tasks, learn multiple knowledge, and adapt to different environments, similar to human capabilities.

From a cooperative perspective, AI facilitates the expansion of IE by attracting diverse participants for collaborative innovation and strengthening relationships through technology integration and data sharing. 
Conversely, from a competitive perspective, AI enhances the efficiency and capabilities of participants, intensifying the competitive relationship and driving IE to accelerate its innovation processes.

\subsubsection{Ultimate Goals in Thinking Space}

The definition and description in Table \ref{Table: Summary of IE's Definition and Description from Physical, Social, and Thinking Spaces} highlight another essential element, which is that the ultimate goal of IE is to achieve value co-creation, reflecting the collective aspirations of participants in the thinking space. 
In terms of value, IE not only seeks economic win-win results but also aims to enhance and enrich relationships. 
The continuous development of both economic and relational value facilitates the attraction of additional funds, technology, and market share, thereby boosting the competitiveness of IE and establishing a virtuous cycle of development.

Beyond the direct value mentioned (explicit associated value), the development goals of IE also encompass indirect value (implicit associated value), specifically dynamic capability and innovation capability (\cite{sun2022impact}). 
The influence of these four dimensions of value is subtle yet critical for the sustainable development and capacity enhancement of IE (\cite{sjodin2019knowledge}). 

Furthermore, using AI will also impact IE's ultimate goals, shaping its direction in the thinking space. 
The complexity of AI makes its decision-making process less transparent to participants and accompanied by uncertainty. 
In fields such as healthcare and finance, where transparency is crucial, such black-box decisions can lead to issues related to trust, compliance, and accountability.

A lack of trust may also result in perceived unfairness in AI decision-making, which could undermine both internal and external cooperative-competitive relationships within IE, posing a potentially significant threat to the overall security and stability of the ecosystem. 
Therefore, With the addition of AI to IE, responsibility capability emerges as the fifth-dimensional goal of IE's development.

\begin{figure*}[ht]
  \centering
  \includegraphics[scale=0.12]{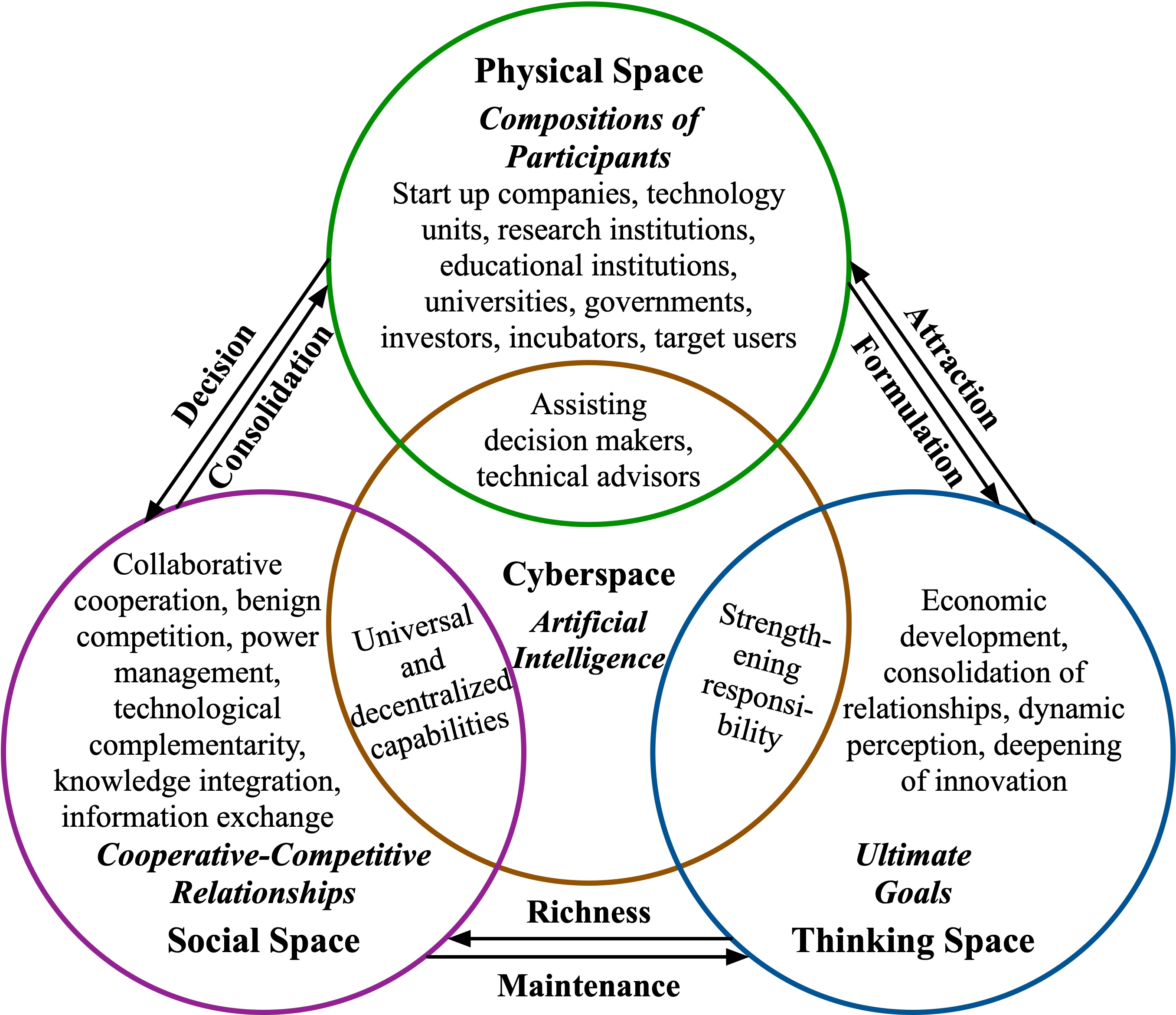}
  \caption{The Impact of AI on IE from Spatial Dimensions}  
  \label{The Impact of AI on IE from Three Dimensions}
  \end{figure*}

Additionally, achieving IE's goals can have a reciprocal effect on the composition of participants. 
By expanding the ecosystem's scale and increasing its diversity, these goals can further strengthen and develop cooperative-competitive relationships, contributing to the ecosystem's sustainable development.
Thus, the compositions of participants, cooperative-competitive relationships, and ultimate goals together form the three essential components of IE.
The impact of AI on these three aspects in IE can also be described in Fig. \ref{The Impact of AI on IE from Three Dimensions}.

Based on the analysis of concepts and definitions related to IE, the paper introduces a description of AIIE.
AIIE is a derivative concept of traditional IE and equally applies to the three essential elements discussed. 
However, unlike traditional IE, AIIE is dominated by AI technologies, which introduce changes and impacts on these elements.

AIIE is a sustainable digital community driven by AI technology, dedicated to providing auxiliary decision-making capabilities, overcoming temporal-spatial barriers in a decentralized manner, and achieving five-dimensional ultimate goals that are competitive, benign, dynamic, innovative, and transparent. In this concept, the emphasis is placed on the role positioning of AI and its unique role in the three spaces corresponding to IE. In AIIE, AI is not only an advanced technology but also the backbone of building a digital, informative, and intelligent ecosystem. In the physical space, AI-assisted decision-making can reduce dependence on professional institutions, expert knowledge, and specialized skills, and enhance the ability to process, analyze, and respond quickly to information. In the social space, AI's decentralized management can further accelerate resource sharing, break down physical temporal-spatial limitations, and heterogeneous information barriers. In the thinking space, AIIE is not limited to achieving the goals of traditional IE but also aims to improve transparency to enhance its responsibility capability.

In the above definition, AIIE also highlights two other key points. 
Firstly, the concept defines sustainability, indicating that AI's role will evolve at different periods of IE development. 
Secondly, the ultimate form of AIIE is to create a digital community that fosters a high degree of inclusiveness and recognition for diverse innovation and development.
Although research has begun to propose AIIE and related descriptions, it remains at a relatively general level of description for the ecosystem. In contrast, the paper maps AIIE to four spaces, providing a more comprehensive explanation of its characteristics. To highlight the novelty of the proposed AIIE, the paper compares the concept with descriptions in other papers from six aspects: core definition, AI status, collaborative way, spatial range, resource integration, and target, and summarizes them in Table \ref{Comparison of proposed AIIE with other related concepts}.

\hyphenation{inter-de-pen-dent com-ple-men-tary col-lab-o-ra-tion}
\begin{sidewaystable}
\caption{Comparison of Proposed AIIE With Other Related Concepts}
{
  \renewcommand{\arraystretch}{1.2}
  \begin{tabular}{p{2.5cm}p{4.2cm}p{2cm}p{2.7cm}p{2.5cm}p{3.3cm}p{3.5cm}} \toprule
    Concept & Core definition & AI status & Collaborative way & Spatial range & Resource integration & Target \\ \midrule
    AIIE  (\cite{sun2022impact}) & The interdependent network of participants involved in the development of AI & Participation & Dependent on network relationships & Overcoming geographical barriers & Network configuration resources & Maintain network stability and dynamic development \\
    AIIE  (\cite{ceccagnoli2012cocreation,kuhl2020supporting}) & Integrating complementary resources through open collaboration to improve AI innovation efficiency and effectiveness & Dominant & Proactive open collaboration & Overcoming geographical barriers & Complementary resource integration (such as technology, data, capital) & The efficiency and effectiveness of AI innovation \\
    AIIE  (\cite{sarker2012exploring}) & Realizing the value output of AI innovation through open collaboration and resource sharing among entities & Dominant & Open collaboration for resource contribution & Overcoming geographical barriers & The direct value output of resource sharing & Realize value co-creation \\
    Entrepreneurial IE  (\cite{roundy2024understanding}) & Local actors and factors that coordinate and support AI technology creation within specific geographic regions & Participation & Localized collaboration & Restricted to specific geographic regions & Dependent on regional resources & Promote the implementation of regional AI technology and economic growth \\
    Data ecosystem (\cite{toorajipour2024data}) & A complex network composed of interconnected entities to create, exchange, and utilize data, jointly creating value & Participation & Create, exchange, and utilize data & Overcoming geographical barriers & Data and cross-domain knowledge are strategic assets and critical resources & Realize new forms of data-driven value creation \\
    AI entrepreneurial ecosystem (\cite{roundy2024understanding}) & The location where entrepreneurs and organizations seek to create AI technology within this ecosystem holds significant importance & Participation & Location based attributes and sharing of attribute related resources & A fertile environment for pursuing AI innovation & Financial resources (such as venture capital and financing) and human capital (such as knowledge and research and development) & Emphasize the richness of products, technology, and local environment \\
    AIIE (Ours) & A sustainable data community centered around AI technology & Dominant & AI assisted decision-making & Overcoming geographical barriers & Decentralized sharing & Value co-creation goals that are competitive, benign, dynamic, innovative, and transparent \\ \bottomrule
  \end{tabular}
}
\label{Comparison of proposed AIIE with other related concepts}
\end{sidewaystable}

\subsection{Characteristics of AIIE}

AI, as the core driving force of technological development, exhibits unique and powerful multidimensional characteristics. These characteristics not only reflect the evolution logic of technology itself but also meet the core requirements of modern society for intelligent systems.

The design, implementation, and application of AI depend directly on the form and quality of the data. Compared to traditional single-function models, the heterogeneity of AI enables it to integrate diverse data forms, ensure continuous system optimisation in dynamic environments through its learning capabilities, and avoid knowledge solidification. In addition, adaptability refers to the ability of AI to respond flexibly to uncertainty and boundary-blurring problems, thereby providing opportunities for cross-domain task migration.

The above characteristics are concentrated in the process of AI training and testing. In practical applications, AI has also demonstrated its unique expressive power. The decentralization capability is beginning to receive attention and improve system robustness and privacy security, while risk management mechanisms reduce decision risks through relevant constraints. Furthermore, personalized services reflect AI's precise response to differentiated needs, while efficient resource management capabilities are crucial to achieving sustainable development.

The above seven characteristics support and complement each other, conforming not only to the objective development law of technological evolution but also responding to the expectations of trustworthy, efficient, and humanized AI. These characteristics constitute the core pillars of the AI system's rationality and also serve as the direction for future AI development efforts. Therefore, the paper analyzes the content of relevant literature, summarizes and infers the differences between traditional IE and AIIE in these dimensions, and summarizes them in Table \ref{Table: Comparison of Characteristics between Traditional IE and AIIE}.

\begin{table}
\caption{Comparison of Characteristics Between Traditional IE and AIIE}
{
  \renewcommand{\arraystretch}{1}
  \small
  \begin{tabular}{p{2.6cm}p{6.3cm}p{6.3cm}} \toprule
    Characteristics & Traditional IE & AIIE \\ \midrule
    Heterogeneity & Traditional IE is similar to a cluster, which represents the critical mass necessary for achieving extraordinary competitive success in a specific field and location. Participants in the ecosystem share similar backgrounds, cultures, and other characteristics. Traditional IE consists of Silicon Valley and technology parks dedicated to studying regional competitiveness and economic benefits, demonstrating co-location and regional focus (\cite{porter1998clusters}). & Specific geographic locations no longer define the boundaries of AIIE but by collective functions. Participants exhibit diversity in terms of background, culture, and other aspects. AIIE comprises companies such as Apple and Google, which utilise AI to facilitate efficient information exchange. It is committed to using AI to achieve interconnectivity and demonstrates collaborative attributes. \\
    Sustainability & Traditional IE primarily relies on expert knowledge and experience for learning. While this way offers good transparency, it also incurs high learning costs and significant subjectivity. & The ability of AI to generate or self-generate has become the main driving force for IE's sustainable development (\cite{boyer2020toward}). \\
    Adaptivity & Strategy and process adjustments require manual interventions such as research, analysis, and judgment. Data integration and processing are relatively slow, with a long update cycle. & AI can influence the innovation capability of IE through digital adaptability (perceptual adaptability, social adaptability, and production adaptability) (\cite{gao2025artificial}). \\
    Decentralization & Existing IE typically relies on centralized intermediaries to record behavior, and a highly centralized hierarchical structure can result in unfairness for more dispersed regions to benefit (\cite{mandych2023risk}). & AI can achieve service discovery and recommendation in a distributed manner, ensuring that IE has higher scalability and maintainability (\cite{gao2018dses}). \\
    Risk assessment & Relying on historical data, research findings, and expert experience for analysis and judgment offers high reliability but lacks real-time and dynamic adjustment capabilities, which may be constrained by the limitations of experience. & AI can be adapted to analyze the risk level of collaborative innovation projects and serves as an effective mechanism for detecting and even predicting potential innovation risks in IE (\cite{abreu2018fuzzy}). The continuity of this analysis process, coupled with the involvement of strategic management and the use of detection tools, can identify weaknesses, alleviate decomposition, and implement improvements and updates to IE (\cite{mandych2023risk}). \\
    Personalization & Personalization is somewhat limited, typically offering tailored services based on basic attributes like user profiles and regional distribution, which makes it challenging to address more complex needs. & AI can help AIIE undergo a series of transformations and upgrades tailored to individual needs based on data analysis and personalized interactions (\cite{secundo2024transformative}), adapting to the flexible and changing market demands and meeting the needs of target groups (\cite{zhang2023innovation}). This is key to achieving the digital revolution (\cite{jones2024rewriting}). \\
    Resource allocation & The process tends to be cumbersome, requiring numerous manual operations and approvals, which can be easily constrained by organizational structure and established procedures. & AI can be employed for collaborative collection, sharing, and exploration of a large amount of information, data, and knowledge resources that support IE dynamics (\cite{mercier2020platform}), blurring the boundaries of resource inflow and outflow in IE through an open innovation model (\cite{corrales2024co}). \\
    \bottomrule
  \end{tabular}
}
\label{Table: Comparison of Characteristics between Traditional IE and AIIE}
\end{table}

\emph{Heterogeneity.} With the development of AI, this technology can now achieve low-latency, high-quality semantic hierarchical communication by processing heterogeneous data, such as text, audio, images, and videos (\cite{wang2021risk}). This characteristic facilitates the employ of AI to enhance information sharing, break down language, regional, and cultural barriers to communication, and promote cooperation among participants within the same system. Therefore, AIIE can overcome temporal-spatial barriers, unite participants with diverse backgrounds, and facilitate communication-based on shared goals, thereby enriching regional characteristics.

\emph{Sustainability.} AI is beginning to focus on learning from continuous data streams or task sequences and adaptively developing more complex knowledge and skills over time. This ability to generate and self-generate enables AIIE not only to focus on handling new tasks but also to maintain a memory of history and effectively continue learning when resources are available.

\emph{Adaptivity.} From the perspective of complexity science, if IE is conceptualized as a complex adaptive system, its evolution and development can be more fully understood (\cite{yannier2024ai}). The adaptability of AI enables it to cope with diverse backgrounds, which is challenging to achieve in traditional environments. AIIE will have the ability to dynamically adjust its behavior, models, or strategies in response to environmental changes, data updates, or target requirements. This characteristic enables AIIE to maintain sensitivity and robustness in complex, uncertain, and volatile market environments.

\emph{Decentralization}. Breaking the hierarchical structure of centralized development is conducive to promoting the introduction of technological innovation achievements to various regions, thereby breaking geographical limitations (\cite{sanchez2024design}). According to the research results of graph theory (\cite{shipilov2023user}), the development of IE has three prototypes, namely centralized, adaptive, and decentralized. Decentralization has opened a more equitable door for the development of AI (\cite{montes2019distributed}). By dispersing calculations, data, and decisions among numerous participants, AIIE will achieve significant improvements in reliability, robustness, efficiency, fault tolerance, response time, and other aspects, thereby breaking through the bottleneck of excessive concentration.

\emph{Risk assessment}. With the continuous expansion of IE's scale, the innovation brought by existing resources and the diversity of participants involved has posed management challenges, especially in risk management (\cite{wang2021risk}). AI offers a novel way to risk assessment in IE. Through dynamic risk perception, real-time decision intervention, and predictive governance, AIIE is able to continuously report system stability.

\emph{Personalization.} Personalization involves providing services or products tailored to the needs of the IE target audience (\cite{chandra2022personalization}). However, there is currently a lack of a paradigm to achieve open innovation throughout the life cycle for this characteristic (\cite{zheng2019smart}). AI can enhance the personalized experience of target groups by analyzing massive amounts of data and facilitating interaction and understanding. In the context of constantly evolving personalized demands, the addition of AI enables IE to transform and upgrade towards a personalized mode, thereby better adapting to the changing market environment and the needs of its target group.

\emph{Resource allocation.} IE can be seen as a resource allocation system that helps mitigate the risks associated with IE degradation by creating, allocating, and mobilizing resources (\cite{shi2022unpacking}). By combining various creative approaches, such as dynamic modelling and game theory, AI can provide new ideas for optimal resource allocation and control of IE (\cite{duan2023optimal}). In addition, as data becomes a new key resource, it is of great significance to consider a reasonable, efficient, and complementary resource management model in data-driven AIIE.

It is precisely because AI has unique advantages compared to other technologies that AIIE has stronger capabilities to collect, process, and manage resources compared to traditional IE. This has enabled AIIE to further optimize, supplement, and upgrade traditional IE in terms of compositions of participants, cooperative-competitive relationships, and ultimate goals.

\section{Evolutionary Developments of AIIE}

\cite{moore1993predators, moore1996death} believed that the evolution of IE follows a programmed lifecycle and proposed four stages of IE development, namely birth, expansion, leadership, and self-renewal (or death). \cite{santos2022governance} considered that the development of IE can be seen as four stages: inception, launching, growth, and maturity. According to the research results of \cite{thomas2015processes}, the development of IE is divided into three stages: initiation, motivation, and optimization. When IE is in a leadership state, it is necessary to synchronously consider how to update the system to avoid heading towards death. Therefore, based on the above research results, the paper divides the evolution of AIIE into three stages, namely the startup period, the growth period, and the mature period, and explores the unique role of AI in IE from an evolutionary perspective.

Analyzing the development of the ecosystem from this perspective highlights the importance of dynamic adaptability, where species continuously adjust through natural selection to survive environmental changes. 
Similarly, the compositions of participants, cooperative-competitive relationships, and ultimate goals within AIIE are not static.
They will be constantly adjusted to adapt to changes in the development market and technological environment to achieve sustained progress.
Additionally, due to the differences in development needs at different periods, AIIE's emphasis on evolution will also give it the corresponding characteristics mentioned in Table \ref{Table: Comparison of Characteristics between Traditional IE and AIIE}.

\begin{figure*}[ht]
  \centering
  \rotatebox{90}{\includegraphics[scale=0.09]{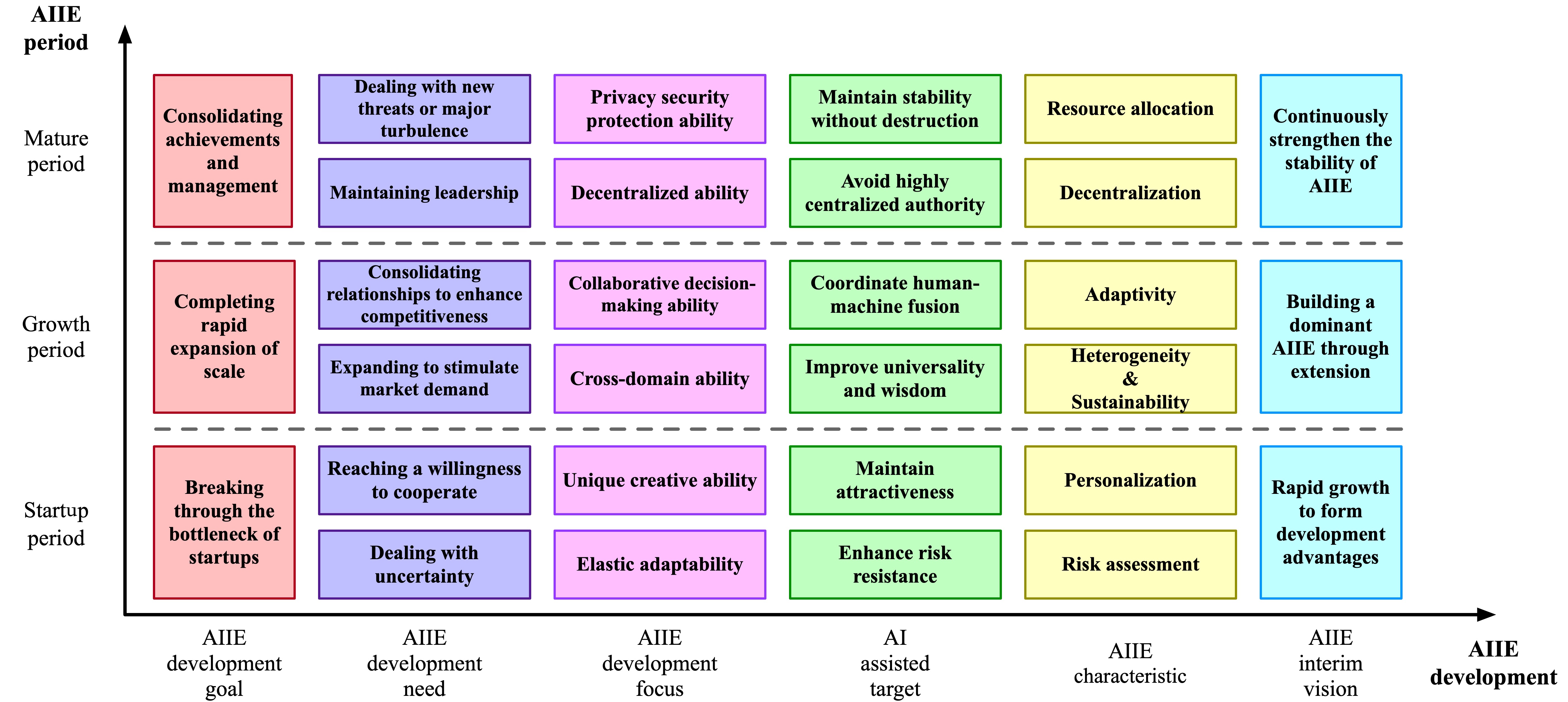}}
  \caption{Three Critical Periods and Priorities of AIIE Development}  
  \label{Three Important Periods and Focus of AIIE Development}
  \end{figure*}

Current studies tend to focus on analyzing the final development form of the ecosystem, emphasizing its relevant attributes and relationships while overlooking the dynamic changes that occur within AIIE throughout its evolutionary process. 
Additionally, the distinct roles AI plays at different periods of this evolution are often neglected. 
To address this gap, the paper examines the development pattern and demand characteristics of AIIE during three critical periods (i.e., startup, growth, and maturity) from an evolutionary perspective. 
This way explores the unique roles and potential impacts of AI at each period. Viewing AI's contributions through the lens of evolutionary theory provides valuable insights into the adaptability, dynamism, and diversity of AIIE, offering a foundation for predicting future development trends. 
The priorities of this analysis across the three critical periods is illustrated in Fig. \ref{Three Important Periods and Focus of AIIE Development}.

\subsection{The Performance of AIIE in the Startup Period}

In the startup period, all participants of IE need to have a common understanding of product and service requirements to smoothly carry out necessary cooperation to achieve common goals (\cite{moore1993predators}). Although AIIE at this period has some uncertainty, it still needs to find a way to form a coherent whole (\cite{markham2010valley}).

Therefore, AIIE's primary development goal during the startup period is to overcome the challenges faced by newcomers, utilize its inherent advantages to achieve a positive willingness to cooperate, and achieve phased common goals for rapid growth.
However, during this period, AIIE may encounter a range of risk factors across multiple dimensions, including policy, finance, law, organizational structure, technology, and operations. 
These risks can undermine development progress and, in some cases, lead to severe setbacks or even catastrophic failure.

At this period, AIIE often operates with relatively limited resources, including funding, talent, and technology, and is typically exposed to an uncertain market environment. 
The combination of scarce resources and fluctuating demand leaves AIIE vulnerable, lacking the capacity for sufficient buffering or adaptive responses to potential risks. 
Additionally, AIIE at this stage may lack a mature organizational structure, management systems, and established processes, which can result in inefficiencies in decision-making, poor internal communication, inadequate emergency management capabilities, and weaknesses in risk identification, assessment, and governance. 
Moreover, this instability can also shake the confidence of participants in cooperation, which is not conducive to forming common insights and achieving benign cooperation in AIIE.
To address these challenges, it is crucial to leverage AI's potential to enhance the resilience and adaptability of the ecosystem, thereby improving its overall stability, robustness, and capacity for long-term growth. 
In a relatively stable environment, AIIE is more conducive to maintaining stability among all parties involved, thereby accelerating the growth of its system towards a common goal. The contribution of AI in elastic settings enables AIIE to have good risk management capabilities and become an effective tool to ensure the stable operation of AIIE.

Furthermore, AIIE at this period is likely to face challenges related to technological innovation and market competition, often lacking the requisite expertise and experience to navigate these pressures effectively. 
This situation is not conducive to the rapid development of AIIE. On the one hand, AIIE, which is in the early period, has a relatively small scale, and these risks can easily lead to its acquisition, merger, or even bankruptcy. On the other hand, the initial cooperation relationship with AIIE is fragile and straightforward. These challenges can lower the confidence of the participants, even leading to withdrawal of investment, cancellation of cooperation, and further exacerbating the development difficulties of AIIE.
To this end, AI can play a pivotal role in boosting innovation, creativity, and imagination within the ecosystem, injecting personalized characteristics into AIIE. 
Through advanced technological research and rapid learning, AIIE can maintain its unique competitive advantage, enabling it to establish itself in a competitive environment and overcome the inherent limitations of early-period development.

\subsubsection{The Elastic Adaptability of AIIE}

The rapid evolution of market and social environments aligns closely with technological advancements, necessitating careful consideration of potential challenges and the formulation of innovative risk management strategies (\cite{endregard2022risk}). 
AIIE needs to be elastic from the beginning of its design, construction, and formation, especially when relying on social technology and in the process of digitalization, integration, and globalization (\cite{pacaux2023calibration,belhadi2024artificial}).
This characteristic encompasses AIIE's ability to swiftly assess, identify, warn, adapt, withstand, recover, monitor, and maintain normal operations amid various internal and external disturbances (\cite{chacon2024cooperative,baryannis2019supply}), whether arising from natural disasters or human-made incidents (\cite{samaei2024using}).

During the startup period of AIIE, it is crucial to quickly overcome the inherent disadvantages of being a newcomer, firmly establish a strong foothold in a complex hazardous environment, and possibly mitigate the negative impacts of various unstable factors (\cite{singh2024augmenting}). 
Frequent disruptive events can easily hinder the normal functioning of AIIE, potentially stalling progress and destabilizing cooperative-competitive relationships (\cite{dohale2024exploring}). 
Furthermore, reliance solely on traditional participants may prove insufficient for effectively managing and sustaining all aspects of the ecosystem.

In comparison to the mature period, AIIE at this period confronts numerous challenges and risks, including political conflicts, anti-globalization sentiments (\cite{wang2022drivers}), resource shortages, economic recessions (\cite{nayal2024role}), market fluctuations, natural disasters, global epidemics (\cite{romero2021towards,zheng2022role}), technological failures, and malicious attacks (\cite{priebe2023sustainability}).

Firstly, AIIE at the startup period typically has limited resources, encompassing funding, manpower, and materials (\cite{le2023linking}). 
Any fluctuations in operations can pose significant threats to AIIE's survival. 
Elasticity allows AIIE to quickly adjust and optimize its operational strategies under resource constraints, ensuring a steady supply of critical resources and minimizing costs (\cite{wong2024artificial}).

Secondly, AIIE in this period often operates within a rapidly changing or emerging market environment, where business demands, collaborative partnerships, technological trends, and even ultimate goals are shifted quickly. 
Elasticity enables AIIE to adapt to these changes, capitalize on new opportunities, and mitigate the risk of being overshadowed by market dynamics.

Thirdly, establishing cooperative-competitive relationships can be particularly challenging for AIIE in its startup period. 
Elasticity aids in building a stable and reliable social network, enhancing trust and boosting the collaborative confidence of other participants. 
Additionally, an elastic way helps maintain service quality, thereby stabilizing relationships by improving the overall reputation.

Finally, as AIIE grows, it must continuously learn and adapt. 
Elasticity facilitates ongoing optimization of development strategies, bolstering overall competitiveness and resilience in the face of challenges, risks, and crises (\cite{wecken2023approach}).

It can be seen that elasticity is essential for startup period AIIE, as it enables AIIE to survive, grow, and expand in environments characterized by limited resources, instability, and intense competition. 
Consequently, AIIE in this period should attach great importance to resilience construction, improve overall robustness, scalability, adaptability, resilience, and stability (\cite{wu2021artificial,raju2024ai}), meet the survival needs in complex and changing environments (\cite{mao2021research}), take anti vulnerability and agility as core design principles (\cite{simpson2021agile}), and be able to maintain synchronization with other participants in unexpected situations (\cite{samadhiya2023influence}).
Ultimately, this characteristic not only supports the sustained development of AIIE (\cite{rane2024artificial}) but also influences the speed and likelihood of returning to a stable state following disruptions (\cite{massel2021integration}).

Currently, there is no standardized definition of elasticity, leading to varied interpretations (\cite{moskalenko2023resilience}). 
Table \ref{Table: Main Description of System Elasticity} outlines the main descriptions of elasticity. 
\cite{dubey2022impact} highlighted that traditional assumptions of commercial supply chains do not apply to humanitarian supply chains, suggesting that elasticity varies across different AIIEs and may influence various developmental periods (\cite{atwani2020ai}). 
While elasticity often operates unnoticed, it serves as a critical sentinel for the entire ecosystem, ensuring adaptability and responsiveness to changing circumstances (\cite{lichti2024change}).

\begin{table}
\caption{Main Descriptions of System Elasticity}
{
  \renewcommand{\arraystretch}{1}
  \small
  \begin{tabular}{p{3cm}p{12.5cm}} \toprule
    Ref & Main description \\ \midrule
    \cite{yu2024resilience} & Elasticity is a system attribute derived from the interactions and coordination of its constituent systems within the architectural framework. \\
    \cite{massel2021integration} & Elasticity refers to a system's ability to return to equilibrium following temporary disturbances. \\
    \cite{mayar2022resilience} & Elasticity is the ability of a system to respond to unexpected changes through passive or active feedback mechanisms, allowing it to restore its state or transition to another suitable form. \\
    \cite{zohuri2022business} & Business elasticity refers to an organization's ability to swiftly adapt to unexpected disruptions, ensuring that workflows continue uninterrupted while safeguarding personnel, resources, assets, and overall operations. \\
    \bottomrule
  \end{tabular}
}
\label{Table: Main Description of System Elasticity}
\end{table}

AI provides a promising solution for building and promoting more resilient ecosystems (\cite{belhadi2022building,schintler2022artificial}), which can improve resilience mechanisms by developing business continuity and perception capabilities (\cite{belhadi2022building,gupta2023influences}), facilitate more effective resource management (\cite{zamani2023artificial}), and provide low-cost, high-quality, high-efficiency, and high-level services (\cite{aliahmadi2022evaluation,ali2022impact}).
In AIIE, the resilience provided by AI is not limited to a single phase but continuously follows through various lifecycle stages, including preparation, response, resistance, recovery, maintenance, and adaptation (\cite{zamani2023artificial,saez2024resilient,maier2024artificial}).
Moreover, alignment between AI (at the technical level) and exploratory learning (at the social level) can strengthen elasticity (\cite{dai2024unveiling}), requiring a balance between optimality and adaptability (\cite{jimenez2013state}). 
In the face of unexpected disruptions, an elastic AIIE should swiftly identify risks and faults, analyze constraints and pressures, reconfigure operations and commands, and activate emergency management and handling (\cite{modgil2022ai}). 
However, the application of AI to enhance elasticity is still in its early stages and represents a relatively new vision (\cite{usigbe2024enhancing,patalas2024integrating}).

\cite{belhadi2022building} proposed an integrated multi-criteria decision-making strategy that employs fuzzy systems, wavelet neural networks, and other methods to support the development of various elastic strategies. 
\cite{dohale2024exploring} introduced a support system based on multi-criteria decision-making, utilizing integrated voting analysis hierarchical processes and Bayesian networks to enhance system elasticity. 
\cite{wong2024artificial} applied partial least squares structural equation modeling and Artificial Neural Network (ANN) for risk management, aiming to improve the system's overall elasticity, reconstruction capabilities, and agility. 
\cite{marshall2017intelligent} developed an adaptive and automated decision system to enhance the inherent elasticity in uncertain environments, primarily within air traffic management. 
\cite{li2024end} designed a data-centric management framework that identifies and evaluates risks in real-time through knowledge accumulation, enabling the deployment of targeted response strategies to enhance overall elasticity. 
\cite{tan2021secure} utilized honeynet methods to bolster Internet of Things (IoT) security and elasticity, incorporating threat detection and situational awareness functions. 
Additionally, evolutionary algorithms, neural network frameworks, and other AI technologies have been validated as practical tools for elastic design, predictive analysis, and risk management, optimizing various decision-making processes (\cite{abaku2024theoretical}).

The above research indicates that AI can offer effective elasticity mechanisms for AIIE. 
However, such a situation does not imply that the elasticity provided by AI is without flaws. 
\cite{singh2024building} suggested that the emergence of unexpected effects is contingent on the transparency afforded to all participants, which is precisely what AI lacks. 
Furthermore, AI may impose limitations on the operational range of AIIE's elasticity, leading to potential constraints (\cite{yu2024resilience}).

\subsubsection{The Unique Creative Ability of AIIE}

A common perception is that AI lacks the ability to think and create (\cite{haase2023artificial}), traits considered uniquely human (\cite{koivisto2023best}). 
However, relying solely on human participants' wisdom and imagination for innovation has become ineffective, particularly in symbiotic ecosystems like AIIE (\cite{gulya2023rebranding}). 
Currently, GAI (or creative AI) serves as a crucial technological foundation for exploring creativity in the era of generative wisdom (\cite{fenwick2023originality}). 
This technology has transformed AI's role from performing data-driven, discriminative tasks to engaging in complex, creative endeavors (\cite{fenwick2023originality}), reshaping the human creativity way (\cite{andrews2023copyright}). 
GAI integrates various advanced technologies (such as machine learning) to generate novel content (such as text, images, and music) by analyzing patterns in training data (\cite{ooi2023potential,mannuru2023artificial,zhou2024generative}). 
When AIIE outputs original and practical results, this situation indirectly enhances its industry attractiveness, fostering collaboration relationships and attracting external support crucial for overcoming startup limitations (\cite{meilin2020study}).
The distinctions between human participants, machines, and disciplines, as well as between formal and informal learning, are increasingly blurring, and GAI will accelerate this trend (\cite{gulya2023rebranding}). 
Moreover, integrating GAI into AIIE's innovation process may help mitigate the cognitive limitations of human participants (\cite{sarica2023innovation}).

ChatGPT and other GAI products enhance search experiences, reshape information generation and presentation, and serve as new entry points for online traffic (\cite{lv2023generative}). 
\cite{guzik2023originality} conducted an exploratory survey on ChatGPT's creative abilities, evaluating this tool through the Torrance Creative Thinking Test. 
The results indicated that ChatGPT demonstrates fluency, flexibility, and originality, suggesting it can elevate and enhance the creativity of human participants. 
\cite{haase2023artificial} found no qualitative difference between the creativity produced by AI and that generated by humans, though the processes for idea generation differ. 
The integration of GAI can expand AIIE's expressive capabilities and reduce barriers to realizing original ideas, such as skill deficiencies in production (\cite{fox2018domesticating}). 
However, since creativity likely stems from multiple sources, it is essential to employ appropriate tools for managing the creativity generated by AI (\cite{botega2020artificial}).

It is important to recognize that GAI primarily functions as a probabilistic model, lacking the ability to reason or understand (\cite{garcia2023new}). 
It relies heavily on extensive existing knowledge and data for training to identify potential patterns or structures (\cite{bozkurt2023generative}). 
Consequently, its outputs may carry risks of plagiarism and/or copyright infringement (\cite{cooper2023examining}). 
\cite{yin2021good} discovered that a Transformer model trained using traditional early stopping criteria often produces repetitive patterns, leading to low output quality and originality. 
In later training stages, the model tends to overfit, resulting in reproductions of input segments. 
\cite{zerouati2024examining} noted that while ChatGPT can enhance creativity by providing new ideas, thoughts, and insights, it may also hinder originality by guiding the creation of derivative or non-original works. 
\cite{majovsky2023artificial} confirmed that current AI systems can generate entirely fabricated scientific papers. 
Additionally, GAI is likely to utilize both human-created and self-generated content for future training, potentially creating a feedback loop between generative data and public data. 
\cite{martinez2023towards} found that when trained on such mixed data, the quality and diversity of generated model output images decreased over time, further indicating that merging AI-generated data could negatively impact future model iterations.

The debate surrounding the protection of AI-generated creative works remains contentious, with three primary perspectives emerging. 
Firstly, some researchers argue that intellectual property laws were created to regulate works produced by humans and Human Intelligence (HI), asserting that AI does not fall within this scope and therefore cannot be considered as a co-author (\cite{mihuailua2023artistic,he2024ai}). 
Secondly, other researchers advocate for rediscovering concepts related to innovation, creativity, and originality, including protections for AI-generated outputs. 
Lastly, some suggest that existing laws, regulations, and orders should be leveraged to navigate the complexities surrounding AI-generated works (\cite{ribeiro2020intellectual}). 
According to the current development trajectory, AI will make the creative output forms of AIIE increasingly complex, so accelerating the formulation of copyright protection laws is an effective compromise.

Determining the ownership and authorship of AI-generated works presents significant challenges (\cite{mazzi2024authorship,eshraghian2020human}), yet it remains a critical issue within copyright law. 
The judicial interpretation of such creativity and works varies, resulting in a lack of unified definitions and considerable ambiguity (\cite{linke2020copyright}). 
Nevertheless, evaluating creativity is essential for assessing whether GAI is protected (\cite{lu2024perspective}).

Moreover, for derivative works (improvements to existing works), the identification rights of their authors and the division of ownerships also need to be taken seriously (\cite{diaz2023fusion}). 
Therefore, establishing clear standards for originality is crucial in determining the legal status of AI-generated content. 
Current theoretical discussions on copyright often explore contrasting views of ``negative dialectics" and ``positive dialectics" (\cite{han2024perspective}) or analyze originality through a fusion of subjective and objective criteria (\cite{lu2024perspective}).

Objectively, AI-generated content shares the characteristics of traditional copyrighted works, fulfilling essential attributes such as appearance and information consumption functions. 
Subjectively, the process by which AI generates content reflects the creative thinking processes of human participants (\cite{lihua2024reflection}). 
The concept of originality as defined in US federal law can still effectively address these new forms of creativity, with the two fundamental requirements of modern copyright law (i.e., tangible medium of expression and a minimum level of creativity) remaining relevant for determining legal protection (\cite{fenwick2023originality}). 
Similarly, EU copyright regulations are currently equally applicable (\cite{hugenholtz2021copyright}).

From a broad perspective, AI creativity aligns with the emotions and intentions of human participants involved in the development of AIIE rather than conflicting with them. 
In fact, AIIE should encourage the establishment of partnerships between human participants and AI, viewing this collaboration as a means to maximize the creative potential of both parties (\cite{mazzone2019art}). 
While AI has been shown to offer new modes of creative expression and participation in AIIE, it has also raised questions among human participants about agency and the role of AI in the creative process (\cite{yusa2022reflections}).

A survey (\cite{chiarella2022investigating}) indicated that when AI claims to have produced abstract artworks, negative biases can emerge compared to human claims. 
\cite{magni2024humans} found that the public sometimes perceives AI-generated products as less creative than those produced by humans. 
This skepticism extends to areas like news headlines, where AI-generated headlines are often viewed as lacking the accuracy of those crafted by humans due to the absence of human motivation and emotions (\cite{longoni2022news}). 
Additionally, \cite{babl2023generative} demonstrated that while ChatGPT-generated paper abstracts may not contain obvious errors, they can cite non-existent papers, raising concerns about academic integrity (\cite{farrelly2023generative}).
Therefore, continuous efforts are still needed to enhance organizational collaboration and public awareness so that AI-driven creative models can be widely accepted and recognized both inside and outside AIIE (\cite{gaffar2024copyright}).

\subsection{The Performance of AIIE in the Growth Period}

After overcoming the bottleneck in the startup period, AIIE has accumulated certain development advantages. By setting up AI-based resilience mechanisms, AIIE can cope with uncertainty and has gained experience in risk resistance. Through the unique creativity of AI, AIIE has gained attractiveness in the industry and provided a technological foundation for achieving common goals. To achieve further expansion and better meet the evolving market demands and application scenarios, AIIE needs to strengthen its capabilities to meet the business needs of continuous expansion.

\cite{moore1993predators} pointed out that the first task of IE in the expansion period (that is, the growth period) is to expand to new application fields. When AIIE's market share in the original field is close to saturation, the growth space will be limited. It needs to find increments by entering new fields and expanding into the new world to stimulate market demand. If there is an over-reliance on a single field, it not only tends to enhance the risk of returns but is also not conducive to meeting the continuously upgraded expectations of users. The AI development layout has subverted many traditional industries, injecting new vitality into development. With the development of AI capabilities, its degree of universalization has been significantly improved, providing a solid technical foundation for cross-domain communication and cooperation. Therefore, the cross-domain capability of AI has become a crucial attribute in the development of AIIE during its growth period, serving as the technological foundation for enhancing AIIE's heterogeneity and sustainability.

However, cross-domain communication can also lead to new growth issues. When multiple cross-domain AIIEs target the same application domain, there will be more intense competition. This competitive situation not only affects the participants' technology, business models, etc. but also includes the competition for resources (such as talent competition and supply chain control competition). Therefore, AIIE leaders must be able to maintain good and strong relationships with numerous participants. As a technological leader, AI often has black-box characteristics. This characteristic makes AI decisions not entirely trustworthy and even questionable. In the AIIE growth period, the precision, reliability, and transparency of AI decisions will be more conducive to consolidating cooperative-competitive relationships within AIIE. Therefore, it is necessary to consider collaborative decision-making strategies in AIIE during this period. Through this interaction mode, not only can the transparency of AIIE decision output be improved, but AIIE can also have good adaptability across domains.

\subsubsection{The Cross-Domain Ability of AIIE}

As AIIE rapidly expands, cross-domain communication and collaboration become essential development tasks. 
To support this growth period, AI needs to possess interdisciplinary capabilities to help establish a dominant AIIE. 
While the advancements in AI for specific purposes are noteworthy, achieving AGI remains a distant goal (\cite{fjelland2020general}). 
AGI refers to AI that matches or exceeds HI (\cite{clune2020ai}), capable of adapting in situations with limited knowledge and resources (\cite{wang2019defining}).
It has the ability to reason contextually, choose perspectives, process ambiguous information (\cite{roli2022organisms}), creatively design solutions (\cite{moruzzi2020artificial,hein2018can}), perceive context (\cite{kejriwal2021essential}), and find general answers to complex problems. 
The concept of AGI envisions a future intelligent agent that could surpass even the most talented human minds (\cite{sublime2024ai}).

However, there is currently no consensus on the specific AGI functionalities, standardized AGI terminology (\cite{mclean2023risks,maruyama2020conditions}), or normalized AGI construction methods (\cite{rosa2016framework}), leading to significant disparities between AGI's theoretical framework and the actual needs of AIIE (\cite{kumpulainen2022artificial}).

Artificial Narrow Intelligence (ANI) refers to an inductive logic system that performs specific intelligent behaviors in defined environments (\cite{hunter2024we}), while strong AI corresponds to what is known as AGI (\cite{gobble2019road}). 
This comparison highlights the differences between AGI and ANI. 
Although AGI can be considered a weak AI, it approaches strong AI due to its universality (\cite{fjelland2020general}).

Most current AI applications are designed for limited, specific tasks. 
However, as AIIE evolves, many scenarios require AGI, which can address various tasks without specialized design, thereby minimizing redundancy (\cite{triguero2024general,fitzgerald20202020,togelius2016general}). 
Ideally, AGI would learn from unlabeled training data and reduce the need for human interaction (\cite{szegedy2020promising}). 
A notable example is ChatGPT, which exemplifies the potential capabilities of AGI, demonstrating exceptional performance across many natural language processing tasks and achieving impressive results without specialized training (\cite{amin2023will}).

The development of AGI relies heavily on the quality and accessibility of data provided by participants in AIIE. 
However, the fragmentation and variability of data hinder effective utilization. 
In resource-constrained situations, general capabilities will inevitably be limited (\cite{hernandez2021general}). 
The evolution of AGI involves a transitional phase from ``learning from data" to ``learning what data to learn from", shifting the focus from effectively training models to acquiring and utilizing task-related data (\cite{jiang2023general}).

The decentralization of data ownership and the diversity of data formats in AIIE often result in inefficient data retrieval and processing, significantly impeding AGI's advancement in research and application. 
To address these challenges, \cite{he2024opendatalab} developed a standardized multimodal and multi-format data representation platform utilizing next-generation AI dataset description language. 
This platform aims to bridge the gaps between various data sources and fulfill unified data processing requirements.

Most AI relies on understanding the principles of intelligence as a prerequisite. 
There are two general ways of developing AGI, namely for computer science and neuroscience.
The inherent differences in formulas and encoding schemes between these ways, along with their reliance on incompatible platforms, have significantly hindered AGI's progress.

To overcome these challenges, it is necessary to develop a universal platform that can support models based on computer science, as well as algorithms inspired by neuroscience, that is, to build a so-called neural computer.
This platform could be trained to foster autonomous intelligence and AGI (\cite{huang2017imitating}). 
\cite{pei2019towards} designed the Tianjic chip, which integrates these two methodologies, creating a hybrid, collaborative platform.
Currently, the development of AI chips increasingly focuses on achieving neural computing with low power consumption and cost-effectiveness while comparing the generality, performance, robustness, and scalability of these chip solutions against human-like intelligence capabilities to accelerate AGI development (\cite{james2021and}).

\cite{williams2020model} discovered that minimal reducible functions can represent all cognitive models, offering a pathway for developing AGI with general problem-solving abilities. 
However, most current AI systems exhibit only a single cognitive capability. 
To address this limitation and establish a foundation for AGI, \cite{fei2022towards} proposed a basic model pre-trained on a large volume of multimodal data, enabling rapid adaptation to various downstream cognitive tasks. 
Additionally, \cite{williams2020defining} suggested creating a model library that would provide AGI with a diverse selection of models, allowing for better adaptation to different application scenarios.

The prevailing way to build AGI involves simulating the development of biological human brains, empowering AGI through interdisciplinary frameworks such as genetic algorithms (\cite{clune2019ai,bauer2023using}). 
Inspired by the brain's evolution, \cite{nadji2020brain} created the artificial brain that exhibits cognitive abilities through processes akin to biological brain growth. 
This method incorporates key characteristics of biological brains, including spiking neurons, neural plasticity, and neuron pruning, allowing the artificial brain to learn, retain information, and self-improve through genetic algorithms. 
\cite{lu2018brain} proposed developing an intelligent learning model named ``Brain Intelligence", which generates novel insights about events without direct experience, leveraging artificial life with imaginative capabilities. 
\cite{pontes2019conceptual} introduced a conceptual framework for biomimetic parallel and distributed learning, designing a spiking neural network-based proxy control mechanism that enables self-learning through environmental rewards. 
\cite{mizutani2018whole} focused on creating a whole-brain connectivity architecture aimed at developing a universal biologically plausible AI that can perform various cognitive functions and behaviors similar to the human brain.

In contrast to the previously mentioned methods for building AGI, \cite{latapie2021metamodel} found that meta-learning allows AI to analyze and reason about various learning mechanisms, fostering highly collaborative interoperability. 
This way not only elevates AI's autonomous learning and optimization capabilities but may also offer insights into enhancing human cognition. 
Regardless of how AGI is designed, evaluating the effectiveness of these frameworks cannot rely solely on existing methods for assessing human or non-human animal consciousness. 
Instead, a more reasonable way involves developing universal heuristic evaluation methods that are intuitively sound, theoretically neutral, and scientifically manageable (\cite{shevlin2020general}). 
\cite{maruyama2020conditions} discussed how to test AGI and proposed a novel Turing test centered on the concept of intelligent survival.

Current estimates suggest that AGI could emerge before 2060, with some scholars predicting it as early as 2040 (\cite{salmi2023democratic}). 
While the exact timeline remains uncertain, AGI's development may arise from converging advanced technologies such as big data, deep learning, and quantum computing (\cite{obaid2023machine}). 
Conversely, some hold a negative view, arguing that AGI is theoretically impossible. 
\cite{obaid2023machine} have revisited Hubert Dreyfus's perspective, asserting that computers do not truly engage with the world. 
Additionally, another common viewpoint posits that, due to the constraints of real-time and space, no real-world system will have an undefined generality (\cite{gobble2019road}).

AGI can provide enormous benefits to AIIE, but it also introduces significant risks (\cite{salmon2021putting}). 
These risks include the possibility of AGI operating beyond human participants' control in AIIE, setting unsafe goals, exhibiting poor ethics, perspectives, and value orientation, mismanagement of AGI (\cite{mclean2023risks}), and potentially threatening human survival (\cite{naude2020race,cohen2020asymptotically}). 
Given AGI's advanced cognitive abilities, it's crucial to consider whether it can be regarded as a trustee or governed by contract law (\cite{linarelli2019artificial}). 
To realize AGI's advantages, it is essential to align its objectives with participant values through responsible governance. 
As AGI approaches or even surpasses HI, AIIE must confront complex social and ethical issues related to autonomy, consciousness, and destructiveness (\cite{obaid2023machine}).

\subsubsection{The Collaborative Decision-Making Ability of AIIE}

During the growth period, AIIE will encounter increasingly complex and uncertain environments for innovation, marketing, and survival dynamics, which will pose significant challenges for decision-making, updates, and modifications. 
While human participants, particularly experts, possess valuable domain knowledge and can understand the context needed for effective decisions, they often struggle with collecting, processing, and analyzing large volumes of high-speed data (\cite{zagalsky2021design}). 
Thus, it is essential to effectively integrate AI to enhance and complement human cognition through collaborative decision-making, creating a synergy that addresses the inherent limitations of both (\cite{korteling2021human}).

However, decision-making tasks within AIIE cannot be fully entrusted to AI (\cite{dellermann2021future}), both now and in the foreseeable future. 
Ethically, AI must prioritize human interests in its decision-making processes. 
This principle stems from its fundamental purpose of serving human participants, the necessity of ethical and moral considerations, the importance of fostering fair and inclusive cooperative-competitive relationships, the potential to enhance human capabilities, and the urgency to mitigate growth risks.
Technically, there remain gaps between AI and human participants in areas such as data sources, transparency, understanding, and application effectiveness (\cite{zagalsky2021design,ed2022addressing}). 
AI relies on limited data to learn decision features, establish decision boundaries, and enhance generalization capabilities. 
In the face of complex, ambiguous, and emerging challenges during AIIE's growth period, the absence of data can lead to inaccurate AI decisions. 
Furthermore, the opacity of AI decision-making complicates the explanation of results and limits the ability to formulate effective solution strategies (\cite{johnson2022integrating}).

Moreover, granting AI full authority to make decisions within AIIE, while leaving human participants unaware of the underlying processes or only intervening during significant errors (\cite{pacaux2017designing}), could lead to irreversible losses and consequences, even jeopardizing the stability and development of AIIE. 
Given these limitations, AIIE's decision-making must not entirely exclude human intervention nor overlook the capabilities of AI (\cite{wang2020learning}). 
Therefore, the role of AI in collaborative decision-making should center on human participants (\cite{herrmann2023keeping}), aiming to supplement and enhance human participants' abilities, guided by innovative thinking, and focusing on the long-term goal of sustainable development for AIIE (\cite{zohuri2020artificial}).

Human-machine fusion intelligence represents a third form of intelligence that synergistically combines HI and AI, resulting in capabilities that exceed the sum of both (\cite{yao2023human}). 
This fusion can mitigate subjective biases and cognitive limitations inherent in human participants (\cite{smirnov2021methodology}) while enhancing the transparency of AI decision-making processes (\cite{fan2021disaster}). 
It also facilitates improved situational assessment, decision-making, and coordination among various stakeholders. 
By fostering collaboration between human participants and AI, this way effectively manages and controls the complex relationships within rapidly expanding AIIE (\cite{hafez2020human,dzobo2020integrating}). 
Related concepts include super AI (\cite{zohuri2020artificial}), cooperative intelligence (\cite{sendhoff2020cooperative}), collective human-machine intelligence (\cite{gupta2023fostering}), integrated AI (\cite{rovzanec2023human}), and enhanced intelligence (\cite{yau2021augmented}).

In such AIIE, established human participants must actively engage in the control loop and assess AI's decision-making performance to determine appropriate levels of participation (\cite{abbass2019social}). 
This way allows for the wise allocation of tasks and the definition of roles for both HI and AI (\cite{zhang2024dynamics,trunk2020current}), ultimately fostering an intelligent ecosystem with capabilities for self-organization, self-learning, self-adaptation, and self-evolution (\cite{guo2022crowdhmt}). 
Therefore, AIIE should adopt a people-oriented strategy (\cite{rovzanec2023human,baicun2020human}), designing new decision-making processes (\cite{pacaux2017designing}) that enhance positive interactions while clearly defining the scope of tasks (\cite{bullock2022machine}) that AI is responsible for completing.

In this pattern, the status of human participants is elevated to a supervisory level, with higher-level decisions made through the corresponding human-machine interface only when necessary to override autonomy (\cite{chen2021human}). 
In addition, this pattern can involve HI and continuously align AI with and respect human values (such as rationality, responsibility, and self-esteem) (\cite{sendhoff2020cooperative}), accelerate AIIE growth, promote positive communication, and enhance risk management capabilities. 
\cite{smirnov2021methodology} suggested that ontology methods can improve human-machine interoperability in systems like enterprise management, which require intensive information and knowledge exchange across loosely connected dynamic autonomous domains. 
However, multidimensional ontology integration is essential for decision-making processes that necessitate cross-domain knowledge integration.
Within the loop design pattern of AIIE, the principle of control asserts that AI serves human participants, adhering to the rationality of ethical values (\cite{dong2020research}). 
If AI ever surpasses HI in certain contexts, implementing termination functions, such as ``termination switches", will serve as a safeguard to protect human dominance in AIIE (\cite{ishii2019comparative}).

In the decision-making process of the human-machine fusion intelligence pattern, AIIE must address two key issues. 
On the one hand, it needs to navigate the potential conflicts between AI and HI during decision-making. 
It is worth paying attention to the tasks and conditions under which decisions can be safely entrusted to AI for processing, but in which situations human participants need to intervene and reasonably resolve conflicts that may arise between the two in decision-making (\cite{korteling2021human}).

In cyber security, \cite{zhang2022artificial} introduced the Human in the Loop Cyber Security Model (HLCSM), which characteristics two primary components, i.e., the machine detection module and the manual intervention module. 
The machine detection module leads the process, while the manual intervention module is supportive. 
To facilitate collaboration and clarify control ownership, HLCSM includes a confidence module that assesses AI decisions, only transferring decision-making power to human participants when inconsistencies arise. 
Once participants address these conflicts, the results are incorporated into the knowledge base to enhance AI capabilities.

Similarly, \cite{ye2022parallel} proposed a parallel cognitive modeling paradigm designed to continuously learn and iterate, reducing conflicts in human-machine decision-making. 
This paradigm comprises three elements, i.e., descriptive cognition based on artificial cognitive systems, predictive cognition derived from computational thinking experiments, and guided cognition built upon behavioral interaction guidance. 
Preliminary experiments in specific scenarios indicated that parallel cognition effectively guides human behavior, enhancing human-machine collaboration in complex systems.

On the other hand, the emotional empowerment of AIIE in the human-machine fusion process presents another crucial issue. 
While emotions, moods, and personalities can lead to biased decisions among human participants, achieving a higher level of human-machine symbiosis requires intelligent cognitive abilities in AIIE that are similar to or even surpass those of humans. 
Such an ecosystem should not only be able to engage in human-like thinking, reasoning, and analysis, but also be able to display human-like motivations, emotions, and personalities, and capture human cognitive performance in skills, summarization, and memory (\cite{sun2020potential}).
From both chemistry and sociology perspectives, the dual role of human-like characteristics, such as emotions, in ecosystem development warrants careful consideration (\cite{magni2024digital}). 

\cite{filieri2022customer} argued that designing human-machine fusion intelligence aims to leverage AI's scalability and efficiency while mitigating its limitations, such as poor explanatory power and restricted emotional range. 
\cite{dong2020research} suggested that AI essentially serves as an analogy to HI, indicating that current AI can only simulate, replace, or extend specific aspects of HI.
For AI to possess higher ideological consciousness in the future, it should ideally have the capability to simulate or even replace various personalized abilities, including imagination, emotions, intuition, potential, and tacit knowledge. 
This situation also raises important issues for AIIE regarding whether to grant ``privacy rights" to such sentient AI and what the scope of these rights should entail (\cite{blodgett2018future}).

\subsection{The Performance of AIIE in the Mature Period}

After expansionary growth, AIIE's position and reputation have stabilized within the industry, with its scale, relationships, and processes reaching a relatively stable state. AIIE has begun to enter a mature period of development, which is the highest form in evolutionary perspective. At this period, AIIE should not slack off in managing the entire massive ecosystem but should build on existing resources, clarify a clear vision for future development, and continuously achieve self-renewal rather than decline or even death. Therefore, AIIE's work in the mature period is a self-optimizing process.

AI is the core technology in AIIE, therefore its leadership is needed to complete the control and management of the entire AIIE. However, as the ecosystem continues to upgrade and expand in scale, efficient management methods must be sought. Centralized management can consolidate AI across a single or a few participants for unified management and deployment. This way can achieve efficient integration of resources, facilitate collaboration through standardization and unified norms, and centralize security and risk control. However, as the scale of AIIE continues to expand, the drawbacks of centralized management also begin to emerge, including single point of failure risk, power monopoly, reduced flexibility, and resource imbalance. Therefore, with the distributed, decentralized, and hierarchical management capabilities of AI, it is necessary to carefully examine the mature AIIE management model to maintain the decentralized leadership of the entire ecosystem.

Regardless of the development's period, AIIE needs to process and analyze digital information, which serves as both the fuel and driving force for the rapid development of AI. When significant changes and turbulence occur in the information environment, such as regulatory changes, demographic shifts, and cyber threats, AIIE must provide timely feedback and effective handling of these new challenges to maintain overall stability. Especially in the mature period, participants will engage in extensive information exchange, and information privacy and security will become particularly important. Therefore, AI has a responsibility to protect the confidentiality and security of information, maintaining the stability of the entire ecosystem and achieving efficient management of precious resources.

\subsubsection{The Decentralized Ability of AIIE}

Traditionally, AI has adopted a centralized architecture, methods, and learning mode (\cite{phansalkar2019decentralizing}), so all aspects of its lifecycle (development, training, testing, deployment, etc.) are implemented in a centralized manner (\cite{gupta2020decentralization}), promoting the development and progress of centralized communication and decision-making (\cite{vergne2020decentralized}).
However, in the mature period of AIIE, monopolizing core AI technology by a few large oligopolies can stifle the growth of non-oligopolistic participants. 
This inequitable environment not only limits AI's potential but may also undermine the stability of AIIE (\cite{montes2019distributed}). 
Therefore, it is essential to eliminate centralized control and address the challenges inherent in AI (\cite{singh2020blockiotintelligence}).

Decentralized AI, marked by distributed, permissionless, and democratic access, is regarded as the next wave of development (\cite{gupta2020decentralization,kersic2024review,adel2022decentralizing}). 
This technology signifies a paradigm shift in AI, aimed at redistributing control and resources within AIIE to broader stakeholders (\cite{aaron2024review}). 
To facilitate this transition from centralization, technologies such as blockchain and cryptography (\cite{gupta2020decentralization}) will play a crucial role in advancing AI within a distributed management framework, ultimately achieving the decentralization and democratization of AI through symbiotic relationships (\cite{beniiche2021way}).

As an open, decentralized, and distributed public ledger, blockchain has emerged as an ideal solution for enhancing the security and privacy of distributed AI. 
It offers a decentralized and immutable data storage pattern (\cite{saleh2024blockchain}), enabling cryptocurrencies to securely record changes without the need for third-party verification (\cite{krittanawong2022artificial}).
The rapid advancement of this technology has also led to the rise of AI-driven Decentralized Autonomous Organizations (DAOs), sometimes referred to as Decentralized Autonomous Companies (DACs) (\cite{yadlapalli2019artificially}). 
In this organizational structure, management and operational rules are typically encoded as smart contracts on the blockchain, allowing for autonomous operation without centralized control or third-party intervention, thereby managing participant interactions (\cite{wang2019decentralized}).

As a new economic organizational mode, existing ecosystem governance theories and methods are not fully applicable to DAOs (\cite{wang2019decentralized}). 
Therefore, it is necessary for AIIE to consider transforming its management pattern after maturity, by utilizing high-performance computing and AI that can securely manage AIIE data exchange (\cite{krittanawong2022artificial}), building a distributed ledger technology based on blockchain to create a blueprint for distributed, decentralized, and democratic development, opening the door to more equitable development (\cite{montes2019distributed}), providing richer virtual tools in practice (such as intelligent information analysis and remote collaboration), and helping to promote research on related methods (such as dispersing data permissions through distributed methods) (\cite{krittanawong2022artificial}).
AIIE's distributed characteristics are expected to challenge traditional hierarchical management patterns, significantly reducing communication, management, and collaboration costs (\cite{wang2019decentralized}).

In such a pattern, significant changes will occur in the management and autonomy of AIIE. 
Firstly, the distributed framework allows AI to move away from centralized learning, enhancing its capabilities through dispersed data. 
This decentralization encompasses not only geographical distribution but also the ownership, format, governance, and access permissions among participants, thereby improving AIIE's flexibility, reliability, and security.

Secondly, adjustments to ecosystem architecture, major decision revisions, and other related activities will no longer be determined solely by participating oligarchs. 
Instead, these tasks will be divided into smaller components and processed by dedicated AI programs or experienced participants.

Thirdly, distributed ledgers will facilitate the maintenance and exchange of information among participants, enhancing trust and security through attributes such as immutability, transparency, provenance, and tamper resistance (\cite{lobo2020convergence}). 
Any transaction that alters the state of information will be recorded in the distributed ledger, making it easily trackable. 
Access to information will be restricted to authorized entities, ensuring that even management-level participants cannot modify any records (\cite{rana2022blockchain}).

Under the influence of decentralized AI, AIIE can implement new management measures for participants to strengthen cooperative-competitive relationships. 
Utilizing the token mechanism of blockchain, AIIE can transform into a token-based economy. 
In this pattern, participants need to obtain tokens and prove ownership (\cite{de2017neuron}) to enhance efficiency, inclusiveness, and participation (\cite{basly2024artificial}). 
This way facilitates broader cooperation and competition, fostering more frequent creative interactions among qualified participants who can uncover new insights within AIIE (\cite{lobo2020convergence}).

\cite{teerapittayanon2019daimon} developed a decentralized AI network called DaiMoN, which maintains a decentralized ledger that can only be appended to record key information. 
This network incentivizes collaboration by rewarding contributing peers with encrypted tokens. 
\cite{ponomarev2017multi} proposed a globally distributed system that employs rewards based on smart contracts provided by Ethereum, allowing multiple interacting agents to communicate with each other and environments without reliance on a central authority. 
\cite{eisses2018effect} introduced an open, decentralized network called Effect, which offers technical services in the AI market, replacing several existing services without charging fees. 
This low-barrier entry promotes rapid industry growth and drives enterprise development through tokens.

However, the decentralized management pattern may also present new challenges for the development of AIIE. 
Firstly, the heterogeneity, asynchrony, and latency of information can create conflicts among participants, potentially impacting collaboration and business efficiency (\cite{alirezaie2024decentralized}). 
Secondly, diverse stakeholders may have varying requirements for management concerning resource utilization, collaboration, and privacy (\cite{singh2017internet}). 
Lastly, unpredictable logical and code vulnerabilities within management mechanisms could significantly damage AIIE (\cite{ding2022novel}).

\subsubsection{The Privacy Security Protection Ability of AIIE}

The development of AIIE through AI has both positive and negative implications (\cite{cheng2022good}). 
Frequent interactions among participants have led to many previously private activities becoming public, making some private information easily accessible (\cite{sigalas2021artificial}).
At different periods of development, AI needs to integrate, process, and clean up information in AIIE, which may pose unique threats to privacy security protection and impact ethical processing and safeguarding of information (\cite{shahriar2023survey}).
Therefore, it is essential to prioritize privacy and security throughout the entire lifecycle of AIIE, rather than focusing solely on specific development periods.

In the mature period of AIIE, the cooperative-competitive relationships are relatively stable, and the ultimate goal of value co-creation is well-defined. 
However, the complexity of the ecosystem necessitates frequent information exchanges among participants, imposing higher demands on privacy and security. 
Even though AIIE can run relatively stably, AI inherently possesses vulnerabilities and risks, creating opportunities for potential attacks and malicious manipulation (\cite{villegas2023toward}). 
As a result, discussions surrounding privacy and security have received attention (\cite{zhang2021ethics}), becoming crucial not only within academic communities but also for maintaining AIIE's overall stability. 
Additionally, variations in AI will influence the decisions regarding privacy security protections, ultimately impacting the future development of AIIE (\cite{martin2024artificial}).

In recent years, AI has become increasingly vulnerable to advanced and complex hacker attacks, presenting significant risks of abuse (\cite{muhlhoff2023predictive}). 
Common types of AI attacks include Denial of Service (DoS), double spending attacks, malware infiltration, cheating attacks, low-quality local model training attacks, and reverse attacks. 
Insecure events during AI training are termed poisoning attacks, while those during the inference stage (after training) are called evasion attacks.
Poisoning attacks disrupt the training process by maliciously altering the information flow, whereas evasion attacks employ adversarial examples to interfere with inference (\cite{bae2018security}). 
When confronted with these security challenges, if lacking or even losing control, AI may pose a fundamental threat to the stability of AIIE, often surpassing regulatory standards and ethical guidelines (\cite{alonso2023protecting}), and may even leak privacy in competition (\cite{curzon2021privacy,giordano2022use}), resulting in serious security issues.

The challenges posed by security incidents have prompted collaborative research on adversarial AI, focusing on developing robust models that can adapt to various scenarios and withstand different types of attacks (\cite{oseni2021security,al2023review}). 
From the technology perspective, effective control of AI can ensure its positive contribution to AIIE while minimizing the impact of harmful events that threaten the integrity and efficiency of the ecosystem (\cite{fadi2022survey}). 
From the privacy and security perspective, traditional centralized AI also exhibits drawbacks, as it is not entirely suitable for sensitive data-driven scenarios (\cite{hao2019efficient}). 
In an irreversible disaster, leveraging AIIE's decentralized capabilities is essential for swiftly cutting off connections between the damaged components and the whole, thereby mitigating potential losses.

Even within the same AIIE, some participants may feel dissatisfied with allocating usage rights, resources, and other aspects. 
In such cases, they might intentionally falsify information to alter the balance of power, which poses a serious threat and disruption to AI's decision-making (\cite{cloarec2022privacy}). 
However, if this method is applied positively, it can enhance the security of AIIE. 
For instance, data obfuscation programs can protect digital and/or physical documents containing sensitive information (\cite{martinelli2020enhanced}). 
\cite{selvarajan2023artificial} employed real intrinsic analysis to convert sensitive features into securely encoded data, reducing the potential impact of malicious attacks.

To ensure that AI decisions align with the development style of AIIE and provide personalized services and proactive support, a significant amount of private data needs to be utilized in the training, updating, and optimization processes. 
This requirement can pressure participants to share data beyond their territory (\cite{meurisch2020privacy}). 
As AIIE matures and stabilizes, the loss of collective privacy emerges as a significant issue, undermining participants' freedom and democracy. 
Through experimental data analysis, \cite{majeed2022group} found that the risk of privacy breaches at the group level is notably high when using AI, with vulnerabilities of group privacy potentially increasing alongside the size and capacity of AI data. 
In such a situation, it is essential to evaluate whether participants' privacy information should be treated as a scarce resource, necessitating careful consideration of shared data contributions to AIIE in compliance with relevant regulations (\cite{pournaras2024collective}).

The privacy protection of AIIE encompasses various topics, including identity recognition authentication (\cite{shahriar2023survey,zhang2024artificial}), participant entity resolution, privacy assistance systems, privacy risk modeling (\cite{samtani2021multi}), participant behavior prediction, digital monitoring, behavior change risks, and information privacy laws formulation, with multidimensional characteristics (\cite{saura2022assessing}). 
By applying protective tools in a targeted manner, risks related to AI-specific components in AIIE can be alerted and reduced in advance (\cite{curzon2021privacy}).

Differential privacy, as a promising mathematical model, can to some extent protect privacy and improve the overall security of AIIE (\cite{zhu2020more}).
This technology offers unique characteristics such as privacy safeguards, security, randomization, integration, and stability (\cite{zhu2019applying}). 
For instance, \cite{chen2021holistic} introduced a differential privacy algorithm for managing confidential information in urban clusters, further enhancing security by optimizing the information storage locations within the designed big data analysis-assisted decision privacy scheme.

Traditional encryption methods, such as one-way encryption (\cite{bandi2022towards}), prevent ciphertext information from participating in operations within the ciphertext domain, creating challenges for the extensive data exchange and analysis required in AIIE. 
In contrast, fully homomorphic encryption allows for processing information without decryption, offering a novel way for the privacy security protection of AIIE (\cite{su2024utilization}).
\cite{vizitiu2019privacy} proposed a new scheme to address the barriers to using privacy information. 
This scheme relies on fully homomorphic encryption to encrypt sensitive data without leaking underlying information and allows AI training, updating, and optimization to be performed directly on floating-point numbers while generating relatively small computational overhead.

Federated learning is also a key privacy protection technology (\cite{khalid2023privacy}) that allows participants to collaboratively train AI models without exposing their local private data and reducing resource consumption (\cite{xu2022security,bai2021advancing}). 
However, in applications requiring extensive interactions, competitors can still exploit shared parameters, potentially undermining the normal operation of AIIE and introducing significant privacy and security threats. 
To address this challenge, \cite{hao2019efficient} proposed a privacy-enhanced federated learning scheme. 
Unlike existing methods, this scheme eliminates the need for interactivity, effectively preventing private data leakage even if multiple participants collude.

Participants in AIIE, including governments and legislative bodies, are increasingly focused on regulating AI use through standards, laws, and rules. 
China, Australia, New Zealand, the US, the EU, and so on are actively advancing privacy laws and information protection institutions to safeguard rights related to privacy and security within AIIE. 
While these diverse standards provide some level of protection, there is still a need for comprehensive improvements (\cite{bartlett2021beyond}).

Additionally, current terms, agreements, and regulations often lack clarity, which can diminish AI's effectiveness in addressing AIIE's privacy and security concerns, leading participants to accept certain risks inadvertently (\cite{peltz2020artificial}). 
Therefore, when formulating and establishing relevant standards, it is essential to address critical issues such as the conceptual boundaries of privacy information (\cite{elliott2022ai}), the implications of AI development for the right to know, and the tension between freedom and security in cross-domain information exchange.

The research results further indicated that the implementation and promotion of standards should determine and adhere to an appropriate balance between privacy security protection and technological development, the overall interests of participants and AIIE, and private profits and public interests. 
Additionally, \cite{wang2022privacy} advocated for the development of specialized standards that provide clear definitions and classifications for different scenarios, along with stricter accountability mechanisms to enhance privacy and security governance.

\section{Case Analysis Based on AIIE}

To further demonstrate the effectiveness of the proposed AIIE, starting from enterprise development cases, the paper analyzes the AI's role in different periods of AIIE development from a practical perspective. Owkin is a biotechnology company founded in Paris, France, in 2016, with a focus on the fields of oncology and immunology. The company's vision is to revolutionize the healthcare industry by leveraging AI to deliver personalized solutions for patients. By harnessing the power of AI, Owkin aims to bridge the gap between data-driven research and clinical practice. Therefore, the key factor for Owkin's rapid growth is the continuous innovative use of AI in drug discovery and development. As the scale continues to expand, Owkin's development model shares similarities with AIIE.

\begin{table}
\caption{Examining the Changes Brought by AI-Centered Development to Owkin From a Spatial Perspective}
{
  \renewcommand{\arraystretch}{1}
  \small
  \begin{tabular}{p{3cm}p{6cm}p{6cm}} \toprule
    & \multicolumn{2}{c}{Components} \\ \cmidrule{2-3}
    Spatial perspective & Traditional IE & AIIE (Affected by cyberspace and newly added) \\ \midrule
    Physical space & Cancer centers, hospitals, biology enterprises, technology companies, universities, investment institutions, patients, drug administration, medical treatment institutions & Auxiliary diagnostic tool (RlapsRisk BC and MSIntuit CRC) \\
    Social space & Clinical trial collaboration, strategic cooperation, research collaboration, channel competition, cost competition, resource integration & Decentralized privacy management, secure connection to decentralized medical resources \\
    Thinking space & Accelerating drug development, providing technical services, focusing on specific fields such as oncology, seeking new treatment methods, reducing risks & Bridge the gap between data-driven research and clinical practice, determine personalized diagnosis plans \\
    \bottomrule
  \end{tabular}
}
\label{Examining the changes brought by AI centered development to Owkin from a spatial perspective}
\end{table}

The first part of the case analysis examines the changes that AI-centered development has brought to Owkin from four spatial perspectives, as shown in Table \ref{Examining the changes brought by AI centered development to Owkin from a spatial perspective}. In the physical space, the compositions of participants is not limited to being equivalent to other traditional pharmaceutical companies. In addition to conventional elements such as hospitals, research institutions, investment industries, and target patients, cyberspace provides auxiliary diagnostic tools as a further supplement to physical space participants. In 2022, Owkin announced that two AI-based diagnostic solutions, RlapsRisk BC and MSIntuit CRC, have been approved for employ in Europe. RlapsRisk BC can predict the risk of recurrence in early breast cancer patients by analyzing digital pathological images. MSIntuit CRC diagnostic tool is based on deep learning methods and can be adapted for microsatellite instability screening of routine H\&E slides in colorectal cancer patients (\cite{saillard2023validation}). These tools have become new diagnostic assistants for doctors, allowing them to focus their energy on high-risk populations and screen patients more effectively.

In the social space, the cooperative-competitive relationship around Owkin is not limited to clinical research cooperation and medical resource competition between pharmaceutical companies but will be presented in a more complex network of relationships. Driven by AI, data as an electronic asset brings more complicated relationships to AIIE in medical scenarios. Unlike public information, medical data has strong privacy. Therefore, a secure management mechanism is needed to ensure the absolute security of data storage. Multiple participants may have access to medical data with trade secrets and are unwilling to share it directly, so improving data utilization to promote entrepreneurship is equally essential. Currently, Owkin is building a global research network that utilizes federated learning to bring together data scientists, researchers, doctors, and representatives from pharmaceutical companies on a research platform that ensures data security and privacy. This decentralized way supports collaborative analysis of cross-global datasets, allowing Owkin to skillfully manage complex data relationships effectively. In addition, Owkin also integrates blockchain technology, allowing participants to complete performance tests by calling models without exposing data.

The concept of AI as the core has also changed the ultimate goal of AIIE built by Owkin in the thinking space. Owkin focuses on the fields of oncology and immunology, aiming to put patients at the centre and integrate precision and personalization to build a brighter future for global healthcare. In addition to providing technical services, accelerating drug development, and exploring new treatment methods, Owkin's development philosophy also centres on AI. The Owkin-built AIIE is committed to utilizing AI to identify new treatment methods, reduce risks, accelerate clinical trials, develop diagnostic tools, and support doctors in helping patients find suitable, personalized diagnoses and treatment plans.

From Owkin's actual development performance, decomposing AIIE into cyberspace and examining its impact on the other three spaces is reasonable. It is precisely this subtle influence that has led to significant changes in the three fundamental attributes of AIIE, i.e., compositions of participants, cooperative-competitive relationships, and ultimate goals, and has begun to value and evaluate the status and role of AI as a new technology.

Additionally, the development ecosystem built around Owkin also shares similarities with the conceptual definition of AIIE. On the one hand, AIIE aims to create a sustainable data community driven by AI, which is consistent with Owkin's current development model. The original intention of Owkin's establishment was to adopt AI as the backbone and continuously expand its scale to broaden the intelligent market in the medical field, and this goal remains unchanged. On the other hand, the concept of AIIE also emphasizes the unique role that AI brings in the three spaces. In the physical space, Owkin designed auxiliary tools to help doctors diagnose quickly. In the social space, Owkin has developed a decentralized platform to help participants overcome temporal-spatial barriers to complete collaborative research. In terms of thinking space, Owkin also focuses on the ultimate goal of transparency, aiming to provide patients with trusted, personalized solutions. Through Owkin's development, it can be seen that the description of the AIIE's conceptual definition is reasonable and can serve as a theoretical basis for guiding the development of AIIE.

The second part of the case analysis is based on Owkin's development experience, analyzing the special abilities of AI in the three periods of AIIE's evolution process. In the nearly 10 years since its establishment, AIIE, built around Owkin, has also undergone a transition from the startup period to the growth period and then to the mature period. Affected and challenged by internal and external environments at different periods, the dominant role of AI is also changing accordingly. Based on the functions of AI in various periods, the paper summarizes the actual events experienced by Owkin and summarizes them in Table \ref{The special role provided by AI in different periods of AIIE composed of Owkin}.

\begin{table}
\caption{The Special Role Provided by AI in Different Periods of AIIE Composed of Owkin}
{
  \renewcommand{\arraystretch}{1}
  \small
  \begin{tabular}{p{2cm}p{2.5cm}p{10.5cm}} \toprule
    AIIE period & Main characteristics & Main performance \\ \midrule
    \multirow{2}{*}{Startup period} & Elastic adaptability & In the project with Amgen to predict the probability of suffering from serious cardiovascular disease, Owkin analyzed and compared various AI methods and provided hundreds of different clinical variables for the program (such as age, medical history, smoking status, cholesterol count, renal function, diabetes, etc.), providing the basis for adjustment of specific clinical use. Additionally, the program can also employ the black-box method and rank which variables have the most significant impact on prediction, which helps build a globally applicable elastic program. \\
    & Unique creative ability & At its inception, Owkin has been discovering and publishing innovative methods in precision medicine, including using AI to analyze complex genetic data and identify new biomarkers for cancer treatment, employing U-Net for cancer diagnosis, launching the Cancer Multiscale Spatial Atlas project, adopting GAI to predict patient responses to treatment, building deep learning models to classify mesothelioma, developing AGI for pathology, training large self-supervised Vision Transformer from histopathology slides (\cite{filiot2024phikon}), and designing the MSIntuit tool based on deep learning for screening colorectal cancer patients (\cite{saillard2023validation}). \\
    \multirow{2}{*}{Growth period} & Cross-domain ability & a) Owkin started with oncology treatment and diagnosis, accumulating datasets of patient data in various forms, including images, voice recordings, and handwritten content. Using advanced transfer learning techniques, Owkin continuously optimizes mathematical models and machine learning algorithms to interpret various omics data, visualize data, and patient profile information. b) Using multimodal patient data, combining histology and molecular profiles, Owkin created detailed features of EP2/EP4 biology. \\
    & Collaborative decision-making ability & a) After obtaining approval from the US Food and Drug Administration, Owkin can apply its AI algorithms in clinical trials to identify subgroups of patients who may benefit from specific treatments, paving the way for personalized medicine. b) Medical experts have verified the data-driven AI drug research and development process, and the most promising treatment combination for the success of clinical trials has been determined, which complements the traditional expert-driven methods. \\
    \multirow{2}{*}{Mature period} & Decentralized ability & Owkin's research network bridges data scientists and international medical researchers, helping to establish and train deep learning AI models on large, dispersed datasets without requiring all participants to pool their resources. \\
    & Privacy security protection ability & a) Utilizing a pioneering collaborative AI framework, federated learning, to enable participants to gain valuable insights from isolated, decentralized, and multi-party data sets and train AI models while protecting patient privacy and proprietary data. b) By building a joint system, Owkin aims to collect data from potentially tens of thousands of patients while maintaining personal privacy. c) In collaboration with NVIDIA and King's College London, Owkin experimented with the feasibility of enhancing the security of medical data using differential privacy frameworks. \\
    \bottomrule
  \end{tabular}
}
\label{The special role provided by AI in different periods of AIIE composed of Owkin}
\end{table}

Owkin, which is in its startup period, needs to establish a precise positioning in the complex market, verify the technological feasibility of using AI as its core competitiveness, and establish preliminary business models and cooperative relationships. Both founders of Owkin have a strong research foundation in the fields of AI and biology. To demonstrate the unique role that AI can play in drug development and medical diagnosis, Owkin conducted extensive and innovative research during this period and publicly published its results in top-tier journals and conferences. After constructing mathematical models and algorithms, Owkin objectively analyzed the effectiveness and progressiveness of this technology throughout the entire process, from target discovery to drug development, by utilizing AI to analyze omics data, visual content, and patient information. Based on these research works, Owkin has cleverly avoided the crushing of mature companies under economies of scale and has a competitive advantage in talent attraction and cooperative financing.

For AI-based healthcare systems, resilience is not just a technical requirement but also a clinical and ethical norm. The complexity of medical scenarios, the sensitivity of data, and the high risk of decision-making make it even more crucial for such systems to maintain reliable operation. Owkin became aware of this situation in the startup period and took corresponding measures to improve system resilience. In the joint prediction of cardiovascular disease project with Amgen, Owkin's AI algorithm was able to better identify the patient population with the highest risk of death, heart attack, or stroke in the past three years by using information collected from over 13700 participants in the placebo group of Amgen clinical trials, outperforming conventional clinical methods. However, individual differences can lead to unstable recognition performance. To improve resilience, Owkin first compared multiple AI methods and analyzed their applicability by providing hundreds of different clinical variables. The typical prognostic factors of heart disease, such as age, medical history, smoking status, cholesterol level, and renal function, are the primary dependencies for good algorithm performance. Secondly, Owkin ranked the contribution of these clinical variables to determine their importance in prediction. Finally, the researchers emphasized that these models must be adjusted for clinical use, as many hospitals' electronic health records do not capture as much patient data as experimental data. Through these elastic mechanisms, AI-based systems have better stability in practical clinical applications, laying a solid foundation for the rapid growth of Owkin-based AIIE.

For Owkin, which possesses multiple core AI technologies, these valuable practical experiences serve as the driving force behind its rapid growth. To further expand the scale of AIIE, Owkin utilises AI's cross-domain capabilities to broaden its business scope. On the one hand, Owkin continues to actively establish strategic partnerships with leading medical institutions and pharmaceutical companies, gaining rich data and expertise through cooperation, including genomics, clinical trial data, and real-world information. These multimodal data can break through the limitations of a single dimension and enable AI to understand a more comprehensive feature space in a complementary way. By using AI to uncover potential relationships between data, it can have good universality in cross-domain challenges such as data distribution, device types, and population differences. On the other hand, Owkin has begun utilising advanced technologies, such as transfer learning, to enable AI models to transfer learned information through knowledge reuse and representation sharing, thereby reducing the repetitive work of model training and significantly enhancing cross-domain capabilities. The design and implementation of these tasks are also beneficial for enhancing AIIE's ability to handle heterogeneous information and making it have the characteristic of continuous learning.

AI-based laboratory results cannot be directly popularized. It is necessary to confirm the actual effect in combination with forward-looking clinical experiments to support personalized medical care. If strict clinical validation is skipped and the technology is directly commercialized, it is easy to increase the error rate of AI diagnosis significantly, and when combined with a decision-making black-box, it will increase user distrust. Therefore, Owkin, in its growth phase, is not only limited to developing advanced medical AI technologies but also actively seeks ways to validate the clinical performance of related models. After obtaining approval from the relevant Food and Drug Regulatory Authorities, Owkin began to conduct clinical evaluations of the predictive performance of AI. This measure can utilize the idea of human-machine integration, with the assistance of medical experts for diagnosis, to determine the most reliable treatment plan for this technology and identify the most beneficial patient groups. Through data-driven AI decision-making and expert-driven auxiliary judgment, this fusion way is expected to provide patients with personalized and accurate diagnostic methods. This human-machine integration way is also beneficial in bridging the gap between data-driven research and clinical practice.

Nowadays, Owkin, which has entered a mature period, has gained a certain level of popularity in the industry, and its cooperative-competitive relationships are relatively stable. As of 2024, Owkin has publicly announced its AI pharmaceutical pipeline for the first time and has already raised \$254 million in funding, including investment institutions such as Sanofi, BpiFrance, and Mubadala. In addition to daily management, data, as a precious asset, is also a key focus of this AIIE. To ensure absolute data security, Owkin did not adopt a centralized way and consolidated all data for unified management. This way not only facilitates the concentration of attack targets but also enables the central node to have excessive permissions, leading to data abuse. Therefore, Owkin collaborated with multiple partners to develop a data platform in 2022. This platform is based on blockchain technology, and enterprises with cooperative-competitive relationships can call relevant AI models to calculate results without exposing their data. This decentralized way supports collaborative analysis across global datasets, effectively managing complex and massive medical data.

An essential characteristic of medical data is its strong privacy, which can have a significant impact on data management, technological applications, and legal compliance. With frequent interactions, Owkin needs to pay more attention to effectively managing these data resources within the vast AIIE system, providing privacy and security protection for the entire ecosystem. To protect patient privacy, Owkin has established a global research network based on federated learning. This network enables participants to securely connect decentralised multi-party datasets, facilitating collaborative training of AI models. In Owkin's recent research, it was found that although federated learning can ensure extremely high privacy and security, data can still be reproduced through model inversion. To further enhance security, Owkin collaborated with NVIDIA and King's College London to investigate the feasibility of differential privacy frameworks in improving the security of federated learning. From recent work, it can also be seen that Owkin's focus has shifted towards privacy and security, which are key factors in maintaining the stability of AIIE.

Over the past 10 years, the development of Owkin has shown a good overlap with the AIIE proposed in the paper, both in terms of spatial composition and evolutionary process. On the one hand, Owkin's development demonstrates that an AI-centered ecosystem can be effectively constructed and reach a relatively stable state during growth under the current technological development path. On the other hand, Owkin's example also demonstrates the rationality of the conceptual AIIE proposed in the paper, as AI will have a profound impact on the compositions of participants, cooperative-competitive relationships, and ultimate goals. In addition, the role played by AI varies at different periods of the ecosystem's development, providing a strong basis for controlling AIIE at various periods.

\section{Future Development Direction of AIIE}

AI can leverage its inherent advantages to play a positive role in the evolution of AIIE across various periods of its development. 
However, due to constraints imposed by factors such as technological capabilities, learning methodologies, and training modes, AI also introduces foreseeable risks to the ecosystem's development. 
These risks primarily manifest in four key areas: interpretability, energy conservation, fairness, and harmony. 
Addressing these challenges will be crucial for advancing the future development of AIIE.

\subsubsection{The Interpretability Development of AIIE}

AI is often perceived as a black-box, with changes to its internal configurations primarily driven by new and improved data (\cite{de2019self}). 
This mode makes the technology not well interpretable and transparent, and the persuasiveness of the results will be reduced, even difficult to understand (\cite{mercier2020platform}). 
Furthermore, this characteristic is disadvantageous for ecosystems prone to change, such as concept drift.
Although Explanatory AI (XAI) has gained traction in academic research, most AI solutions continue to operate as black-boxes.

As AIIE evolves, the participants and their cooperative-competitive relationships will continuously adapt to internal and external demands. 
Therefore, enhancing the overall transparency and interpretability of AIIE is crucial for its healthy development. 
A transparent decision-making process allows stakeholders to understand how decisions are made, thereby reducing misunderstandings, enhancing trust, and strengthening collaboration.

Improving transparency and interpretability facilitates tracking, detecting, and addressing potential biases and discrimination, promoting fairness, accuracy, and objectivity in AI decision-making. 
Additionally, these enhancements empower participants to control better, monitor, and analyze AI behavior, ensuring the overall stability, reliability, and security of AIIE operation, enabling the ecosystem to comply with reasonable regulatory requirements, and avoiding potential legal risks. 
This improvement is significant for AIIE, as any dangerous vulnerability or anomaly could seriously affect the ecosystem, even crossing ethical and legal boundaries.

Moreover, increased transparency and interpretability can improve human-machine fusion, thereby enhancing the efficiency and performance of AIIE. 
While there is growing attention to solutions for AI transparency and interpretability, implementing these capabilities in large and complex ecosystems like AIIE remains a significant challenge.

\subsubsection{The Energy Conservation Development of AIIE}

AIIE requires the employ of AI to assist in decision-making, relationship management, resource integration, and other tasks. Although there has been a significant improvement in automation, the energy consumption behind it cannot be ignored. This consumption can be analyzed from two aspects, namely hardware manufacturing consumption and AI operation consumption.

From the perspective of hardware manufacturing, the production of chips that rely on AI computing power requires a large amount of rare earth resources and hydropower energy. For example, the manufacturing process of a single high-end Graphics Processing Unit (GPU) may generate significant carbon emissions. Taking AIIE's decentralized operation as an example, this way requires participants to have corresponding data processing and management capabilities. This requirement has significantly expanded AIIE's demand for hardware while also exacerbating the problem of electronic waste pollution. Additionally, high hardware costs may also make it difficult for small and medium-sized participants to afford operational and maintenance expenses. At the operational level of AI, a single training session for a general language AI model can result in significant energy consumption. The inference stage also faces energy waste due to the accumulation of computing power, which may lead to the construction of a water cooling system and excessive consumption of water resources. However, these AI models are essential components that give AIIE unique creative capabilities.

In this situation, AIIE needs to consider optimizing energy consumption throughout the entire life-cycle of hardware manufacturing and AI operations to achieve sustainable development. In terms of hardware manufacturing, AIIE can reduce its carbon footprint and costs during the manufacturing phase by developing low-power chips, utilizing renewable materials for packaging, and enhancing hardware energy efficiency, thereby providing hardware deployment advantages for small and medium-sized participants. In terms of AI operation, AIIE can innovate through algorithm compression, dynamic computation allocation, and clean energy supply, among other methods, which not only reduces the energy consumption of AI operations but also maintains model performance. This dual innovation, combining hardware and software, is beneficial for achieving the sustainable development of AIIE and constructing an energy-saving, green operating environment.

\subsubsection{The Fair Development of AIIE}
The development of AIIE relies on the support and collaboration of multiple stakeholders, including participants, society, and even international organizations. 
However, if AI exhibits biases, it can produce unfair, extreme, or erroneous results. 
At a mild level, this situation can disrupt AIIE operations and undermine participant confidence. 
At a severe level, it may provoke public outcry and cross moral or legal boundaries.

Despite AI's application in critical sectors such as healthcare, finance, education, and law, many members of the public remain skeptical of its decision-making capabilities, lacking trust in the technology. 
This skepticism significantly impacts the acceptance of AI within AIIE, which is crucial for the technology's wider adoption. 
If stakeholders harbor biases or doubts about AI, its implementation will be hindered, jeopardizing the harmonious development of AIIE.

Continued biased decision-making by AI can lead to uneven innovation resource allocation, which can dampen the innovation enthusiasm of participants and even cause the breakdown of relevant cooperative-competitive relationships, seriously affecting the overall effectiveness and ultimate goals of AIIE.
Moreover, as an essential part of social development, unfair outputs from AIIE can result in social dissatisfaction and conflict, hindering the establishment of a fair, open, and inclusive innovation environment. 
Such conditions are detrimental to the long-term sustainable development of AIIE and could threaten societal and national stability.

Therefore, it is necessary to eliminate bias against AI as much as possible to ensure the outputs' correctness and fairness and promote the smooth emergence, development, maturity, and prosperity of AIIE. 
Through various measures such as data diversity, model consistency, method interpretability, and audit diversity, AI fairness and bias may be ensured, further promoting the sustainable development and universal acceptance of AIIE.

\subsubsection{The Harmonious Development of AIIE}

The emergence and development of AIIE have significantly changed to the job market, presenting both new challenges and opportunities. 
The integration of AI, particularly through increased automation, has gradually replaced many traditional roles in industrial engineering, such as risk assessors, data collectors, and technical evaluators. 
This trend poses a severe unemployment risk for participants in these positions, especially those with general skill levels who may find it challenging to adapt to the new collaborative demands of working alongside AI in AIIE.

This shift in demand requires participants to continuously upgrade their skills to thrive in the evolving work atmosphere and environment of AIIE. 
However, for some participants, this may represent a daunting challenge. 
The development of AIIE may further exacerbate social inequality. 
The technological dividend may only be concentrated in the hands of a few participants, while most participants may be forced to terminate their collaboration with AIIE due to technological changes, and even face serious social problems such as regional economic imbalances and unemployment waves.

In light of these challenges, AIIE is responsible for promoting the upgrading and transformation of industrial structures, enhancing support for emerging AI technologies, and providing participants with more opportunities to demonstrate their value. 
Additionally, AIIE has the obligation to collaborate with governments and institutions to formulate relevant laws and regulations, establish ethical standards, and regulate AI application behaviors. 
By fostering a harmonious coexistence between AI and human society, AIIE can contribute to creating a fair and sustainable employment landscape.
Only by actively promoting this harmonious development can AIIE gain widespread recognition and realize its potential for innovation, ultimately making significant contributions to individuals, society, nations, and the global community.

\section{Conclusion}

This paper provided a systematic overview of AI's unique role and impact on the development of IE, with the aim of offering guidance and support to researchers in related fields. 
Firstly, the paper summarized the concept of IE and its related terms and introduced the concept of AIIE from a spatial perspective. 
Secondly, the paper analyzed AIIE's uniqueness compared to traditional IE in seven dimensions, emphasizing AI's significant influence on this ecosystem. 
Thirdly, from an evolutionary perspective, the paper examined the distinct roles that AI plays in AIIE development across three critical periods, highlighting AI's transformative impact. 
Subsequently, the paper adopts the development trajectory of the medical company Owkin to analyze and verify the feasibility, effectiveness, and rationality of the conceptual definition of AIIE, as well as the unique role of AI in the evolution and development of AIIE, drawing on relevant examples.
Finally, the paper identified the current limitations of AI and outlined the potential challenges facing the further development of AIIE, which can guide future research directions and priorities. 
In summary, AI's role in advancing IE is continuously expanding, giving rise to a new paradigm of AIIE. 
However, researchers must remain vigilant to the multidimensional risks associated with AIIE's development.

\section*{Acknowledgment}
This work was funded by XXXX (Grant No. XXXXXXXX).

\bibliographystyle{unsrtnat}
\bibliography{ref}  






\end{document}